\documentclass[lettersize,journal]{IEEEtran}
\usepackage{amsmath,amsfonts}
\usepackage{algorithmic}
\usepackage{algorithm}
\usepackage{array}
\usepackage[caption=false,font=normalsize,labelfont=sf,textfont=sf]{subfig}
\usepackage{textcomp}
\usepackage{stfloats}
\usepackage{url}
\usepackage{verbatim}
\usepackage{graphicx}
\usepackage{cite}

\usepackage{xcolor}
\usepackage{caption}
\usepackage{multirow}

\begin{document}

\title{Epidemiology-informed Graph Neural Network for Heterogeneity-aware Epidemic Forecasting}

\author{Yufan Zheng, Wei Jiang,~\IEEEmembership{Student Member,~IEEE,} Tong Chen,~\IEEEmembership{Member,~IEEE}, Alexander Zhou, Choujun Zhan, Quoc Viet Hung Nguyen,~\IEEEmembership{Senior Member,~IEEE,}  Hongzhi Yin,~\IEEEmembership{Senior Member,~IEEE}
\thanks{Yufan Zheng is with the University of Canberra, Canberra, Australia (e-mail: zhjpre@gmail.com).}
\thanks{Wei Jiang, Hongzhi Yin, and Tong Chen are with the University of Queensland, Brisbane, Australia (e-mail: weijiang@uq.edu.au; h.yin1@uq.edu.au; tong.chen@uq.edu.au).}
\thanks{Alexander Zhou is with the Hong Kong Polytechnic University, Kowloon Tong, Hong Kong SAR (e-mail: alexander.zhou@polyu.edu.hk).}
\thanks{Nguyen Quoc Viet Hung is with the Griffith University, Brisbane, Australia (e-mail: henry.nguyen@griffith.edu.au).}
\thanks{Choujun Zhan is with the South China Normal University, Guangzhou, China (e-mail: zchoujun2@gmail.com).}
}

        




\markboth{Journal of \LaTeX\ Class Files,~Vol.~14, No.~8, August~2021}%
{Shell \MakeLowercase{\textit{et al.}}: A Sample Article Using IEEEtran.cls for IEEE Journals}


\maketitle

\begin{abstract}
Among various spatio-temporal prediction tasks, epidemic forecasting plays a critical role in public health management. Recent studies have demonstrated the strong potential of spatio-temporal graph neural networks (STGNNs) in extracting heterogeneous spatio-temporal patterns for epidemic forecasting. 
However, most of these methods bear an over-simplified assumption that two locations (e.g., cities) with similar observed features in previous time steps will develop similar infection numbers in the future. In fact, for any epidemic disease, there exists strong heterogeneity of its intrinsic evolution mechanisms across both geolocation and time, which can eventually lead to diverged infection numbers in two ``similar'' locations.
However, capturing such mechanistic heterogeneity is non-trivial due to the existence of numerous influencing factors like medical resource accessibility, virus mutations, mobility patterns, etc., most of which are spatio-temporal yet unreachable or even unobservable. 
To address this challenge, we propose a Heterogeneous Epidemic-Aware Transmission Graph Neural Network (HeatGNN), a novel epidemic forecasting framework. By binding the epidemiology mechanistic model into a Graph Neural Network, HeatGNN learns epidemiology-informed location embeddings of different locations that reflect their own transmission mechanisms over time. 
On top of the conventional spatio-temporal module, a heterogeneous transmission graph network is designed to uncover the mechanistic heterogeneity among locations, where the time-varying mechanistic affinity graphs are computed on-the-fly and provide additional predictive signals. Experiments\footnote{Code is available at https://anonymous.4open.science/r/HeatGNN-14DB.} on four benchmark datasets have revealed that HeatGNN outperforms various strong baselines.
\end{abstract}

\begin{IEEEkeywords}
Spatio-temporal Graphs, Predictive Analytics, Epidemic Forecasting.
\end{IEEEkeywords}

\section{Introduction}
\IEEEPARstart{T}{he} outbreak and spread of epidemics are massive disasters facing global human society, which can cause a substantial number of deaths or irreversible physiological damage to humans \cite{proal2021long} as well as significant economic losses across various industries worldwide \cite{lenzen2020global}. Recent global outbreaks, such as COVID-19, have placed unprecedented pressure on public health systems and decision-makers \cite{adhikari2020epidemiology}. In response, there has been a growing interest in developing accurate and effective epidemic forecasting models \cite{deng2020cola,zhan2021random}. These models can support the implementation of public health interventions \cite{zhan2023modeling}, optimize resource allocation \cite{chen2023prediction}, and enhance public awareness of early prevention and control strategies \cite{saha2020epidemic}, thereby helping to prevent and control the onset and rapid spread of epidemics. 

\begin{figure*}[htbp]
\centering
\includegraphics[width=0.8\textwidth]{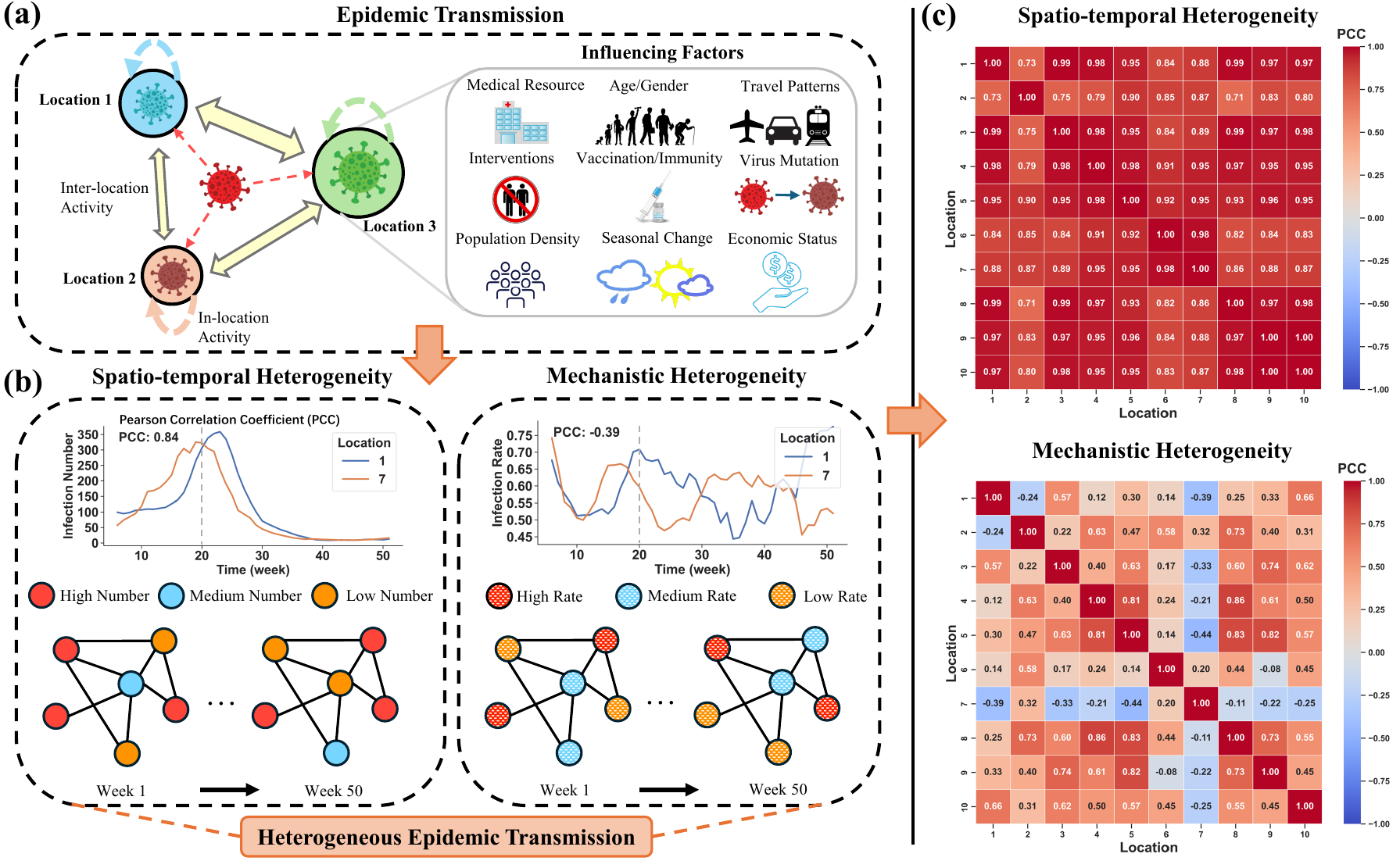}
\caption{Spatio-temporal and mechanistic heterogeneity between different locations in epidemic transmission. (a) Epidemic transmission with different influencing factors. (b) Example of heterogeneous epidemic transmission between two locations. (c) The analyses of two heterogeneities on a dataset (US-Regions). The detailed prior experiment is provided in Section \ref{sec:prior}.\label{fig:prior}}
\vspace{-6mm}
\end{figure*}

In a nutshell, epidemic forecasting predicts the infection numbers at geographically connected regions in the future horizon based on existing records. Traditional epidemic models, like the Susceptible-Infected-Recovered (SIR) model, rely on mechanistic approaches and location networks to simulate multi-regional transmission for epidemic forecasting \cite{kermack1927contribution,zhan2021identifying}. Despite advantageous explainability and applicability, these models exhibit limited capacity to capture real-world complexities.  
As such, with the rapid development of machine learning, methods represented by spatio-temporal graph neural networks (STGNNs) have been used to learn spatial and temporal dependencies in epidemic transmission simultaneously \cite{zhan2021random,deng2020cola,xie2022epignn,qiu2024msgnn,panagopoulos2021transfer,hy2022temporal,nguyen2023predicting}. However, many STGNNs are purely data-driven methods that neglect the underlying epidemic transmission mechanisms, making them prone to overfitting historical data and being misled by spurious features. To address the limitations of both sides, epidemiological machine learning models \cite{wang2019defsi,wan2024epidemiology,rodriguez2023einns} have been proposed by incorporating mechanistic laws into a machine learning backbone to enhance the model's generalizability and robustness. This trend also extends to STGNNs, where the time- and location-sensitive parameters in epidemic models are predicted to support dynamic epidemic forecasting \cite{liu2023epidemiology,cao2022mepognn}.

Generally, a key to epidemic forecasting is heterogeneity discovery, such that similar spatio-temporal patterns among locations can be aligned to enhance predictions while dissimilar ones are filtered out. This also explains the popularity of STGNNs for this task owing to their capability of learning the semantic affinity of connected locations within the topology of the graph. Meanwhile, existing models commonly use an over-simplified assumption that such heterogeneity only exists within observed features, such as past infection number, population, and demographic statistics of a region. However, 
in reality, epidemic heterogeneity holds for not only spatio-temporal features but also the underlying intrinsic epidemic transmission mechanism. In this paper, we refer to these two types of heterogeneity as 
\textit{\textbf{spatio-temporal heterogeneity}} and \textit{\textbf{mechanistic heterogeneity}}, respectively. For illustration, we provide an example in Figure \ref{fig:prior}. In a nutshell, while the spatio-temporal heterogeneity refers to the differences among cities in observed features (i.e., what they are now), such as geographic location and time-dependent factors (e.g., infection numbers)., like infection numbers, the mechanistic heterogeneity highlights the evolutionary trajectories of each location (i.e., how they will change), such as infection rates (that can be described by a well-defined SIR model). It is worth noting that the two heterogeneity types are not mutually substitutable but rather complementary. For instance, though the two locations in Figure \ref{fig:prior}(b) have similar patterns in their infection numbers until the 20th week, the numbers have moved toward different directions in the following two weeks because of the distinct dynamics in the two locations' infection rates. In the real epidemic transmission process, the two heterogeneity signals can vary greatly, as depicted by the Pearson Correlation Coefficient (PCC) across locations from a sampled time step (Figure \ref{fig:prior}(c)). Hence, if only the spatio-temporal heterogeneity is considered in the STGNN, the model is prone to making inaccurate predictions in the long term. 

To this end, bringing awareness of the mechanistic heterogeneity to the prediction model appears to be a beneficial move. However, learning such heterogeneity incurs non-trivial challenges. The first challenge lies in effectively understanding and mimicking the underlying intrinsic transmission mechanisms, which are jointly determined by a complex suite of factors. As shown in Figure \ref{fig:prior}(a), those influencing factors \cite{miller2020disease,zhan2023modeling,liu2021role,davies2020age,kwok2020herd,saha2020epidemic,gibbs2020changing} include (but are not limited to) public health resources, policy interventions, climates, individual or group susceptibility, virus mutations, and so on, many of which are not linked to the epidemic records and are even unobservable. Consequently, this renders explicit and precise modeling of the intrinsic epidemic transmission mechanism infeasible. 
At the same time, the second challenge is the modeling of the similarity/dissimilarity across the epidemic transmission mechanisms of distinct regions. The root cause is the lack of effective measures to represent and quantify the mechanistic heterogeneity across locations, preventing it from providing more predictive signals into epidemic forecasting. 

To address these challenges, we propose an innovative framework for heterogeneous epidemic transmission forecasting modeling: \textbf{H}eterogeneous \textbf{E}pidemic-\textbf{A}ware \textbf{T}ransmission \textbf{G}raph \textbf{N}eural \textbf{N}etwork (\textbf{HeatGNN}). This framework aims to improve epidemic forecasting performance by simultaneously learning spatio-temporal and mechanistic heterogeneity. Firstly, we introduce epidemiology-informed embedding learning to encapsulate the mechanistic patterns into each location's representation. To bypass the intricacy and unavailability of all influencing factors, instead of simulating the explicit mechanistic process of epidemic transmission, we parameterize the time-varying Susceptible-Infected-Recovered (SIR) model for each location with its spatio-temporal graph embedding. By aligning the behaviors of the implicitly parameterized time-varying SIR model and the explicitly derived one, the epidemiology-informed embedding can be effectively optimized.
Secondly, to quantify and take full advantage of the mechanistic heterogeneity across locations, we propose to learn the time-varying mechanistic affinity graphs. Each mechanistic affinity graph is constructed based on the pairwise similarity between two epidemiology-informed location embeddings, where a sparsification approach is in place to remove noisy connections and ensure efficiency. With a backbone STGNN, the spatio-temporal and mechanistic heterogeneities are captured by respectively modeling the geographical and mechanistic affinity graphs across time steps to facilitate heterogeneity-aware epidemic forecasting. In summary, our main contributions are as follows:
\begin{itemize}
    \item In the context of epidemic forecasting, we characterize two types of heterogeneity, namely, spatio-temporal heterogeneity and mechanistic heterogeneity. We point out that on top of the commonly used spatio-temporal heterogeneity, mechanistic heterogeneity is a largely overlooked yet crucial predictive signal for predicting future infection numbers. 
    \item We propose HeatGNN, a novel framework for heterogeneous epidemic forecasting that jointly discovers spatio-temporal and mechanistic heterogeneity. By incorporating a well-defined mechanistic model into the STGNN, HeatGNN learns epidemiology-informed location embedding and mechanistic affinity graphs to account for mechanistic heterogeneity across locations to greatly benefit the predictive performance. 
    \item Our framework achieves state-of-the-art performance across four real-world epidemic datasets. Our results demonstrate that HeatGNN has a high interpretability of intrinsic mechanism transmission across locations and good scalability and robustness in different scenarios.
\end{itemize}

\begin{figure*}[thbp]
\centering
\includegraphics[width=0.8\textwidth]{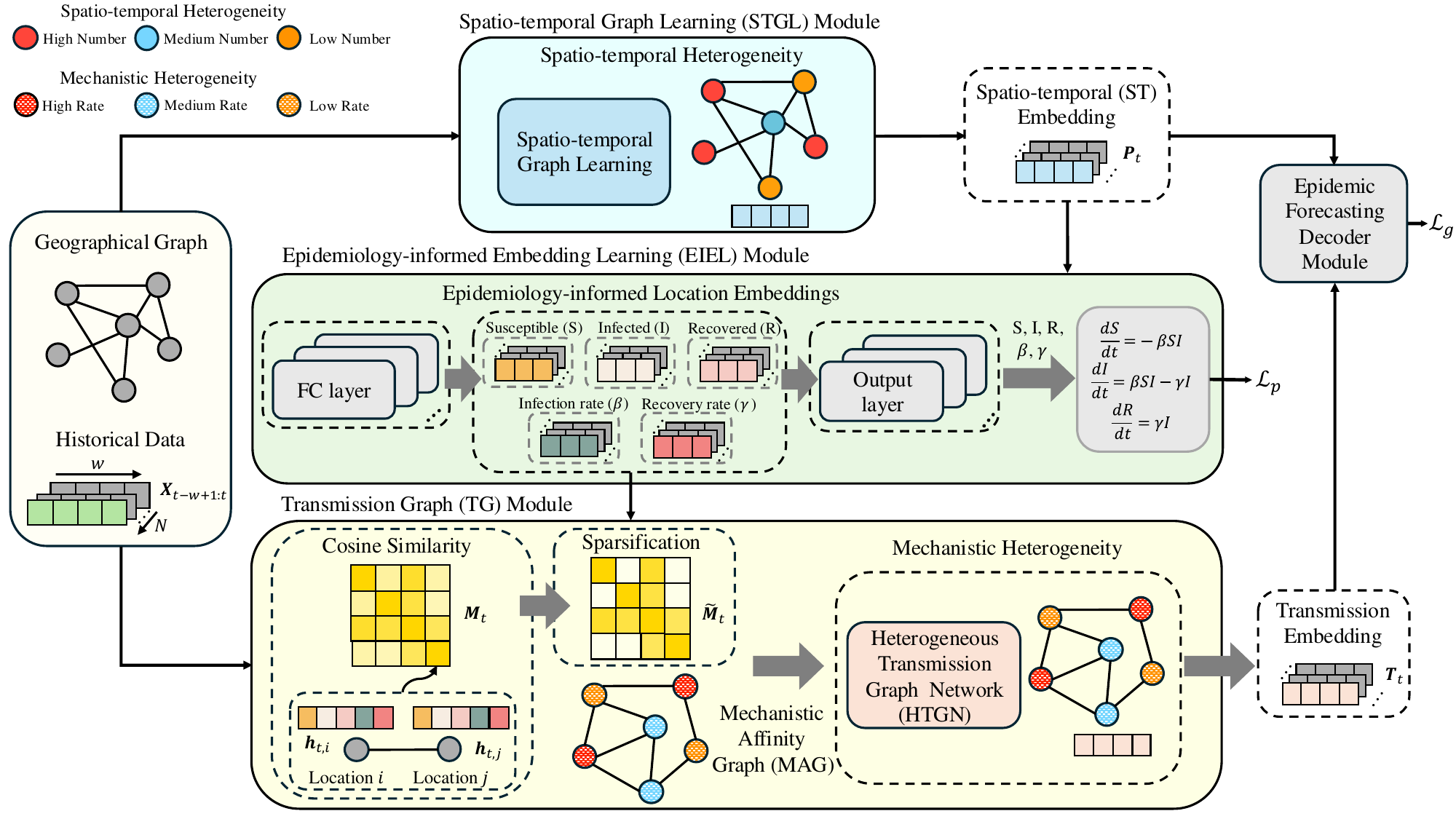}
\caption{The schematic illustration of HeatGNN\label{fig:framework}.}
\vspace{-6mm}
\end{figure*}

\section{Related Work\label{apx:rel_work}}
\textbf{Epidemic Forecasting.} Epidemic forecasting is a critical task in monitoring the spread of epidemics to help public health interventions quickly and avoid large-scale outbreaks \cite{liu2024review}. The time dependence and seasonality of epidemic spread are considered at the beginning of epidemic forecasting. Time series regression models, such as ARIMA \cite{benvenuto2020application} and VAR \cite{shang2021regional}, and machine learning models, such as random forests \cite{zhan2021random}, are widely used to solve this task, but these models cannot capture spatial dependencies and ignore the evolution of epidemic transmission. To further improve the ability of the model to capture temporal dependencies, the RNN and RNN-based models were proposed for this task \cite{wang2020time}. CNN, Transformer, and their adopted models are used to capture multi-scale features, such as short-term and long-term temporal dependencies \cite{wu2020deep}. Meanwhile, the emergence of Graph Neural Network (GNN) provides an effective solution to identifying spatio-temporal patterns in epidemic forecasting. In addition to most studies focusing on the impact of geographic topology \cite{deng2020cola,panagopoulos2021transfer,xie2022epignn}, some studies focus on the distinction between local and global information in epidemic spread \cite{xie2022epignn}, as well as multi-scale information \cite{qiu2024msgnn}. However, these methods mainly focus on the spatio-temporal dependence of epidemic transmission but ignore intrinsic transmission patterns of epidemics. They also lack mechanism models to describe the transmission mechanisms behind epidemics. These limitations make it challenging to provide sufficient information to support public health decision-making.

\textbf{Spatio-temporal Graph Learning.} Spatio-temporal graph learning has emerged as a powerful framework for modeling complex data structures that evolve over both spatial and temporal. GCNs enable end-to-end learning of graph structures by efficiently capturing local features through parameter sharing \cite{kipf2016semi}. GNNs later expanded from static graphs to dynamic graphs to address spatio-temporal modeling needs in dynamic network tasks \cite{zhang2021traffic,deng2020cola}. The development of Graph Convolutional Recurrent Networks allowed dynamic graph updates in time series, capturing dynamic node relationships \cite{seo2018structured}. To integrate temporal and spatial dependencies, STGNNs combine graph and temporal convolutions, effectively learning dynamic interactions \cite{yu2017spatio}. In addition, advancements in STGNNs have incorporated recurrent neural networks \cite{li2017diffusion,cini2023scalable} and Transformer-based attention \cite{zha2024scaling} for more flexible spatio-temporal graph learning. Multiscale STGNNs \cite{li2021multiscale} capture dependencies across different scales, while hypergraph-based STGNNs process high-order interactions \cite{zhao2023dynamic,li2022spatial}. Graph diffusion networks model information propagation \cite{zhang2021traffic}, and heterogeneous STGNNs handle complex spatial-temporal heterogeneity \cite{song2020spatial}.

\textbf{Physics-guided Machine Learning.} The approximations learned by machine learning methods often ignore existing physical knowledge, leading to unexplainable conclusions \cite{wu2024physics}. Therefore, a combination of physics and machine learning has begun to be proposed to solve the problem of lacks of interpretability in machine learning with black-box structures. A commonly used method is PINN \cite{raissi2017physics}, which is based on combining physical residual loss with data loss to distill physical epidemic knowledge into a machine learning framework for constrained learning processes while being able to effectively capture heterogeneous data in different forecasting tasks \cite{jiang2024physics,jiang2024epidemiology,rodriguez2023einns}. Another way to combine physical models and machine learning is neural differential equations (Neural ODEs), which use the neural network as a continuous dynamic process \cite{chen2018neural}. For example, the time-dependent states in the epidemic model are replaced by neural networks \cite{kosma2023neural}.

Some scenarios in epidemic transmission require integrating physical models into machine learning models to improve performance and interpretability. Epidemic models are based on overly simple assumptions and cannot fully understand the epidemic spread \cite{kermack1927contribution}, while machine learning models lack effective mechanisms for explanation. Network-based epidemic models capture transmission patterns between locations but face challenges in large-scale parameter optimization \cite{zhan2021identifying} or data scarcity \cite{gao2023evidence,cao2022mepognn}. To address those problems, some models proposed to use simulated data generated by physical models to solve the problem of insufficient data and provide high-resolution forecasting \cite{wang2019defsi,gao2023evidence}. Meanwhile, machine learning has also begun to estimate epidemic models to solve the parameter explosion 
 \cite{cao2022mepognn,liu2023epidemiology}. Neural ODEs are incorporated into epidemic models to learn the continuous evolution of epidemic spread in different regions \cite{wan2024epidemiology}. The epidemic models were used as causal mechanism models to guide the graph embedding learning process to handle noisy data and missing data better \cite{wang2022causalgnn}. Although physics-guided machine learning has made efforts in epidemic forecasting, previous studies have ignored the mutual influence and dependency of the intrinsic epidemic transmission between locations, which makes it challenging to forecast epidemics breaking out in multiple locations.

\section{Problem Formalization}
In epidemic transmission, we formulate the epidemic forecasting problem as a node-level regression task on graphs. There are $N$ locations of interest in an area; each location (e.g., a region, city, or state) is represented as a node. Given the spatio-temporal observations of infection number: $\boldsymbol{X} = \{\boldsymbol{x}_1,\boldsymbol{x}_2,\ldots, \boldsymbol{x}_T\}^{\top} \in \mathbb{R}^{T \times N}$, where $T$ represents the length of time step (e.g., the infection number or infected population number for $T$ weeks). Each $\boldsymbol{x}_t = \{ x_{t,1},x_{t,2},\ldots,x_{t,N}\}^{\top} \in \mathbb{R}^{N \times 1} (t=1,2,\ldots,T)$ is an $N-$dimensional vector, with $x_{t,i}(i=1,2,\ldots,N)$ being the infection number at the time step $t$ in the location $i$. The epidemic is transmitted in different areas, let $\mathcal{G}=(\mathcal{V},\mathcal{E})$ be a geographical graph, where $\mathcal{V}$ is the nodes set comprising $|\mathcal{V}|=N$ nodes (e.g., regions, cities, or states). The $\mathcal{E} \subseteq \mathcal{V}\times \mathcal{V}$ is the geographic link between these nodes. The adjacency matrix $\mathcal{A} \in \mathbb{R}^{N\times N}$ of the geographical graph is formulated as $\mathcal{A}_{ij} = 1$ if there is an edge $e_{ij} \in \mathcal{E}$, and $\mathcal{A}_{ij}=0$ otherwise. 

The objective of this work is to accurately predict the infection number in each location. Given the historical data and the adjacency matrix of the geographical graph, we can develop a predictive model $f$, which can be formulated as follows:
\begin{equation}
    \boldsymbol{x}_{t+h} = f(\boldsymbol{X}_{t-w+1:t},\mathcal{A}),
\end{equation}
where $\boldsymbol{x}_{t+h}$ is an infection number in a future time step $t+h$ and $h$ represents the forecasting horizon (e.g., the number of time steps to be predicted ahead). $\boldsymbol{X}_{t-w+1:t} = \{\boldsymbol{x}_{t-w+1},\boldsymbol{x}_{t-w+2},\ldots, \boldsymbol{x}_{t}\} \in \mathbb{R}^{w\times N}$ is the historical infection number, $w$ represents the window size of historical input, and $N$ is the number of locations. 

\section{Methodologies}

The proposed framework, as shown in Figure \ref{fig:framework}, consists of four modules: 1) Spatio-temporal Graph Learning (STGL) Module; 2) Epidemiology-informed Embedding Learning (EIEL) Module; 3) Transmission Graph (TG) Module; and 4) Epidemic Forecasting Decoder Module. In the following subsections, we describe each module in more detail.

\subsection{Spatio-temporal Graph Learning Module}
The spread of an epidemic in a city is influenced by nearby cities with similar geographic features, as adjacent locations have similar weather and topographical conditions, leading to comparable epidemic transmission patterns. Conversely, distinct weather and topography can result in varying patterns of spread. These differences lead to the emergence of spatio-temporal heterogeneity in epidemic transmission. To capture spatio-temporal heterogeneity, we propose using the STGL module in epidemic forecasting. The STGL module utilizes historical data to learn dynamic time series variations and geographic interactions, enabling it to capture the spatio-temporal heterogeneity of epidemic transmission across locations. Given the historical infection number $\boldsymbol{X}_{t-w:t}$ and the adjacency matrix $\mathcal{A}$ of geographic topology, the spatio-temporal (ST) embedding $\boldsymbol{P}_t \in \mathbb{R}^{N\times D_{1}}$ with $D_1$-dimensional consists of the embedding $p_{t,i}$ of each location in $N$ locations can be described as:
\begin{equation}
    \boldsymbol{P}_t = f_{\text{ST}}(\mathcal{A}, \boldsymbol{X}_{t-w+1:t}),
\end{equation}
where $f_{\text{ST}}$ is the STGL module. $\boldsymbol{P}_t$ containing spatio-temporal heterogeneity information in $N$ locations. The choice of $f_{\text{ST}}$ is general and flexible, and our HeatGNN can adapt to various spatio-temporal models. Our main innovation lies in the combination of epidemiology mechanistic models and a neural predictor. In our HeatGNN, we use the backbone of EpiGNN as our STGL module \cite{xie2022epignn}.

EpiGNN is a GNN-based model designed for epidemic forecasting, which captures spatio-temporal dependencies by integrating local and global spatial effects. It consists of three key components: multi-scale convolutions, transmission risk encoding, and the region-aware graph learner (RAGL). Multi-scale convolutions extract temporal features across different time scales, capturing both short- and long-term epidemic dynamics. For spatial transmission, EpiGNN encodes local transmission risk (LTR) via network centrality and global transmission risk (GTR) via self-attention, thereby capturing intra-regional and long-distance dependencies, respectively. Leveraging these transmission risk encodings, EpiGNN employs the RAGL to dynamically construct the region correlation graph, integrating local and global dependencies, temporal patterns, and external data. This region correlation graph informs a graph convolutional network (GCN), which aggregates information dynamically from identified neighbors. Finally, EpiGNN integrates GCN outputs with an autoregressive module to effectively capture both nonlinear spatial dependencies and linear temporal trends, thereby improving forecasting accuracy and stability.

\subsection{Epidemiology-informed Embedding Learning Module}
The STGL module only captures the spatio-temporal heterogeneity among different locations, while many previous approaches predict epidemic transmission directly based on the learned ST embedding. However, only capturing the spatio-temporal pattern of the infected population is noisy for epidemic forecasting and lacks the intrinsic epidemic transmission mechanism, which is prone to producing generalization errors in predictions \cite{wan2024epidemiology}. Therefore, we use the epidemiology mechanistic model to enhance the process of learning location transmission dynamics to capture latent patterns and reflect well-defined disease dissemination mechanisms. The time-varying SIR model can effectively capture the dynamic outbreak influenced by various complex factors and is adaptable across different epidemics \cite{bucyibaruta2023discrete,zelenkov2023analysis}. Given a location $i$, we assign the time-varying SIR model as the epidemiology mechanistic model, which can be described as following ordinary differential equations (ODEs):
\begin{equation}
\left\{
\begin{aligned}
\frac{d S_{i}}{d t} &= -\beta_{i} S_{i} I_{i}, \\
\frac{d I_{i}}{d t} &= \beta_{i} S_{i} I_{i} - \gamma_{i} I_{i}, \\
\frac{d R_{i}}{d t} &= \gamma_{i} I_{i},
\end{aligned}
\right.
\label{eq:t-sir}
\end{equation}
where $S_{i}$, $I_{i}$, and $R_{i}$ represent the number of susceptible, infected, and recovered populations in location $i$, respectively. $\beta_{i}$ and $\gamma_{i}$ are the time-dependent parameters, which mean the infection rate of the susceptible population and the recovery rate of the infected population in the location $i$.

Although the time-varying SIR model reflects the intrinsic epidemic transmission mechanism, it requires high computational costs and domain knowledge to predict its large-scale parameters simultaneously. In addition, the variables of the time-varying SIR model are heavily entangled, and it is difficult to focus only on predicting changes in the number of infected populations ($I$) in epidemic forecasting. These challenges motivate us to propose the EIEL module that uses node embeddings to parameterize the five time-varying SIR variables. The EIEL module uses five different multi-layer perceptrons (MLPs) to predict susceptible population number $S$, infected population number $I$, recovered population number $R$, infection rate $\beta$, and recovery rate $\gamma$ in the time-varying SIR model at each location and time step. Each MLP consists of a fully connected (FC) layer with $L$ layers and an output layer. Additionally, to enhance the embedding expressiveness of the EIEL module, we learn epidemiology-informed location embeddings based on the ST embedding $\boldsymbol{P}_t = \{ \boldsymbol{p}_{t,1}, \boldsymbol{p}_{t,2}, \ldots, \boldsymbol{p}_{t,N} \}\in \mathbb{R}^{N\times D_1}$. Taking $\hat{S}_{t+h,i}$ prediction as an example, given the $i$-th location ST embedding $\boldsymbol{p}_{t,i} \in \mathbb{R}^{D_1}$, the susceptible population embedding $\boldsymbol{u}_{i}^{(S)}\in \mathbb{R}^{D_2}$ of epidemiology-informed location embedding in $i$-th location in time step $t$ can be computed by the FC layer $g$ with $L$ layers:
\begin{equation}
    \boldsymbol{u}_{i}^{(S)} = g \left(\boldsymbol{p}_{t,i}; \{\boldsymbol{W}^{l}, \boldsymbol{b}^{l}, \sigma^{l} \}_{l=1}^{L} \right),
\end{equation}
where $\boldsymbol{W}^{l}$ is the weight matrix of the $l$-th hidden layer, $\boldsymbol{b}^{l}$ is its bias vector, and $\sigma^{l}$ is the activation function. Then, the epidemic transmission prediction $\hat{S}_{t+h,i}$ is:
\begin{equation}
    \hat{S}_{t+h,i}=\sigma^{out}(\boldsymbol{W}^{out} \boldsymbol{u}_{ i}^{(S)}+\boldsymbol{b}^{out}),
\end{equation}
where $\boldsymbol{W}^{out}$, $\boldsymbol{b}^{out}$, and $\sigma^{out}(\cdot)$ are the weight matrix, bias vector, and activation function in the output layer of the MLP, respectively. Therefore, the other four parameters ($\hat{I}_{t+h,i}, \hat{R}_{t+h,i}, \hat{\beta}_{t+h,i}, \hat{\gamma}_{t+h,i}$) are predicted using four other MLPs with the same structure, where the parameters of these five MLPs are not shared. Finally, we obtain the epidemiology-informed location embedding $\boldsymbol{h}_{t,i} \in \mathcal{R}^{5\times D_2}$ for location $i$ in time step $t$, which contains five parameter embeddings of time-varying SIR. Moreover, variants of epidemiology mechanistic models such as SEIR or SIRD can be integrated into HeatGNN by modifying the MLPs in EIEL to align with model states and parameters. However, these extensions require additional data, which is often unavailable in mainstream datasets. In Section \ref{sec:optimization}, we introduce the optimization to ensure that the epidemiology-informed location embedding is consistent with the intrinsic epidemic transmission mechanism.

\subsection{Transmission Graph Module}
Although the EIEL effectively captures the intrinsic epidemic transmission mechanisms of individual locations, it does not account for the interactive relationships between them. To address this, we propose a time-varying mechanistic affinity graph (MAG) that can efficiently characterize the mechanistic heterogeneity across locations. Unlike the geographical graph, the MAG provides a comprehensive representation of how transmission mechanisms vary over time and across different locations, reflecting the mechanistic dependencies that influence epidemic transmission. Moreover, to fully capture the complex heterogeneity by leveraging the mechanistic dependencies within the MAG, we propose the Heterogeneous Transmission Graph Network (HTGN), which encodes the transmission (dis)similarity between locations into their embeddings.

\textbf{Mechanistic Affinity Graph.}
The MAG $\widetilde{\boldsymbol{M}}_{t} \in \mathbb{R}^{N\times N}$ is calculated by the cosine similarity function $\phi$ between the epidemiology-informed location embeddings of different locations. The mechanistic similarity between locations is defined as follows:
\begin{equation}
    m_{ij} = \phi(\boldsymbol{h}_{t,i}, \boldsymbol{h}_{t,j}),
\end{equation}
where the mechanistic similarity $m_{ij} \in \boldsymbol{M}_{t}$ represents the mechanistic dependence between locations $i$ and $j$ based on epidemiology-informed location embeddings ($\boldsymbol{h}_{t,i}$ and $\boldsymbol{h}_{t,j}$) of two locations at time step $t$. $\boldsymbol{M}_{t}$ is a symmetric matrix. Then, we normalize the similarity between different locations:
\begin{equation}
    \widetilde{\boldsymbol{M}}_{t} = \boldsymbol{D}_t^{-\frac{1}{2}} \boldsymbol{M}_{t} \boldsymbol{D}_t^{-\frac{1}{2}},
\end{equation}
where $\boldsymbol{D}_{t}$ is the degree matrix of $\boldsymbol{M}_{t}$. To focus on key mechanistic dependence, reduce redundant information, and improve the computational efficiency of GNN, we design an indicator function to sparsify the MAG $\widetilde{\boldsymbol{M}}_{t} \in [0,1]$.
This sparsification is controlled by a threshold $\delta$, applied to each normalized similarity $\widetilde{m}_{ij} \in \widetilde{\boldsymbol{M}}_{t}$:
\begin{equation}
    \widetilde{m}_{ij}= \begin{cases}\widetilde{m}_{ij}, & \text { if } \widetilde{m}_{ij} \geq \delta \\ 0, & \text { otherwise }\end{cases}.
\end{equation}

\textbf{Heterogeneous Transmission Graph Network.}
The HTGN integrates MAG with the architecture of Spatio-Temporal Graph Convolutional Networks (STGCN) \cite{yu2017spatio}. The HTGN consists of two Transmission-Temporal convolutional (TT-Conv) blocks. Each TT-Conv includes two temporal gated convolution layers and one transmission graph convolution. Given the historical infection number $\boldsymbol{X}_{t-w+1:t}$ and MAG $\widetilde{\boldsymbol{M}}_{t} \in \mathbb{R}^{N\times N}$ in time step $t$, we learn the transmission embedding $\boldsymbol{C}_{t} \in \mathbb{R}^{N\times D_3}$ of mechanistic heterogeneity by two TT-Conv blocks. For the input $\boldsymbol{v}^l$ of TT-Conv block $l$, the output $\boldsymbol{v}^{l+1}$ is computed by:
\begin{equation}
\boldsymbol{v}^{l+1}=\Gamma_1^l *_{\tau} \operatorname{ReLU}\left(\Theta^l *_{\mathcal{G}}\left(\Gamma_0^l *_{\tau} \boldsymbol{v}^l\right)\right),
\end{equation}
where $\Gamma_0^l$ and $\Gamma_1^l$ are upper and lower temporal kernels in block $l$, respectively. $\Theta^l$ is the spectral kernel of the graph convolution in block $l$. $*_{\tau}$ and $*_{\mathcal{G}}$ are temporal and graph convolution operators, respectively. $\operatorname{ReLU}(\cdot)$ represents the activation function in block $l$. The input of the first TT-Conv is $\boldsymbol{v}^{(0)} = \boldsymbol{X}_{t-w+1:t}^{\top} \in \mathbb{R}^{N \times w}$. Then, we learn the transmission embedding $\boldsymbol{C}_t = \{ \boldsymbol{c}_{t,1}, \boldsymbol{c}_{t,2}, \ldots, \boldsymbol{c}_{t,N} \} \in \mathbb{R}^{N\times D_3}$ from two TT-Conv blocks.

\subsection{Epidemic Forecasting Decoder Module}
Finally, by simultaneously capturing the spatio-temporal heterogeneity and mechanistic heterogeneity across different locations, our framework applies an FC layer to predict the infection number, providing the epidemic forecasting result $\hat{\boldsymbol{Y}}_{t+h}$ in time step $t+h$:
\begin{equation}
    \hat{\boldsymbol{Y}}_{t+h} = \boldsymbol{W}[\boldsymbol{P}_{t};\boldsymbol{C}_{t}]+\boldsymbol{b},
\end{equation}
where $\boldsymbol{W}$ and $\boldsymbol{b}$ are the weight matrix and bias vector, respectively, $\boldsymbol{P}_{t}$ is the ST embedding learned from the STGL module and $\boldsymbol{C}_{t}$ is the transmission embedding learned from the TG module.

\subsection{Optimization\label{sec:optimization}}
To ensure the epidemiology-informed location embeddings are consistent with the mechanism of the intrinsic epidemic transmission, we use the idea from the physics-informed neural networks (PINN) \cite{raissi2017physics} to propose a mechanistic loss to inform the learning process in the EIEL module. The physics loss incorporates the underlying physical laws of the intrinsic epidemic transmission mechanism into the EIEL module. Specifically, our objective is to solve the time-varying SIR model using five MLPs by minimizing the multi-objective mechanistic loss $\mathcal{L}_{p}$. Given the prediction of five variables ($\hat{S}_{t+h,i}$,  $\hat{I}_{t+h,i}$, $\hat{R}_{t+h,i}$, $\hat{\beta}_{t+h,i}$, and $\hat{\gamma}_{t+h,i}$) of the time-varying SIR model, the loss $\mathcal{L}_{p}$ can be described as follows:
\begin{equation}
\begin{gathered}
    \mathcal{L}_{d}(\hat{I}_{t+h,i}) = \frac{1}{N} \sum_{i=1}^{N} |\hat{I}_{t+h,i}-x_{t+h,i} |,\\
    \mathcal{L}_{o}(\hat{S}_{t+h,i},\hat{I}_{t+h,i},\hat{\beta}_{t+h,i},\hat{\gamma}_{t+h,i}) \\
    =\frac{1}{N} \sum_{i=1}^{N} | \frac{d\hat{I}_{t+h,i}}{dt} -(\hat{\beta}_{t+h,i} \hat{S}_{t+h,i} \hat{I}_{t+h,i} - \hat{\gamma}_{t+h,i} \hat{I}_{t+h,i})|,\\
    \mathcal{L}_{p} = \mathcal{L}_{d} (\hat{I}_{t+h,i}) + \mathcal{L}_{o}(\hat{S}_{t+h,i},\hat{I}_{t+h,i},\hat{\beta}_{t+h,i},\hat{\gamma}_{t+h,i}),
\end{gathered}
\end{equation}
where $x_{t+h,i}$ is the ground truth. $\mathcal{L}_{d}$ represents the error between predicted values and ground truth, which uses mean absolute error (MAE) as a loss measure. $\mathcal{L}_{o}$ is the time derivatives residual between the neural network output and the time-varying SIR model. We use $\mathcal{L}_{d}$ to constrain the output of MLPs to be consistent with the real-world epidemic transmission. As the epidemiology-informed embeddings enable the quantification of mechanistic heterogeneity among locations, $\mathcal{L}_{o}$ puts a constraint on learning the infection trend defined by the epidemic model, i.e., the infection number’s derivative w.r.t. time $\frac{d\hat{I}_{t+h,i}}{dt}$. To achieve this, $\mathcal{L}_{o}$ approximates the second sub-equation in Equation \ref{eq:t-sir} with estimated epidemic model parameters. Guided by $\mathcal{L}_{o}$, HeatGNN can more effectively learn the epidemic transmission mechanism at each location, thereby facilitating the quantification of mechanistic heterogeneity with the subsequent dynamic MAG. Moreover, the constraints are not only the $I$ but also $S$, $\beta$, $\gamma$, and $R$, which are indirectly constrained via the loss $\mathcal{L}_{o}$.

In our framework, we use the forecasting from the epidemic forecasting decoder module as the final forecasting result and use the MAE loss to compare it with the ground truth.
\begin{equation}
    \mathcal{L}_{g} = \frac{1}{N} \sum_{i=1}^{N} | \hat{y}_{t+h,i} - x_{t+h,i}|,
\end{equation}
where $\hat{y}_{t+h,i} \in \hat{\boldsymbol{Y}}_{t+h}$ is the output of HeatGNN and $x_{t+h,i}$ is the ground truth value. Finally, the main objective function $\mathcal{L}$ of HeatGNN is given as follows:
\begin{equation}
    \mathcal{L} = \mathcal{L}_{g} + \lambda\mathcal{L}_{p},
\end{equation}
where the term $\mathcal{L}_{g}$ is the forecasting error, and the term $\mathcal{L}_{p}$ is used as a regularizer of the EIEL module to enhance the ability of this module to represent the intrinsic epidemic transmission mechanism. The loss weight $\lambda$ is used to balance loss terms.

\section{Experiments}

In the experiment, we evaluate HeatGNN by addressing the following research questions (RQs): \textbf{RQ1}: Does HeatGNN outperform the existing state-of-the-art epidemic forecasting models? \textbf{RQ2}: What is the impact of each core component in HeatGNN? \textbf{RQ3}: What is the performance of HeatGNN with different hyperparameters? \textbf{RQ4}: Can the mechanistic similarity reveal intrinsic epidemic transmission across locations? \textbf{RQ5}: Does HeatGNN exhibit robust scalability and efficient inference performance? \textbf{RQ6}: Does the HeatGNN robustness under noisy and incomplete data scenarios?

\subsection{Experimental Settings}

\textbf{Datasets and Metrics.} We evaluate the performance of our HeatGNN using four real-world datasets with multiple wave outbreaks following previous works \cite{deng2020cola,xie2022epignn}: Japan-Prefectures, US-States, US-Regions, and Australia-COVID:
\begin{itemize}
    \item Japan-Prefectures: This dataset comprises weekly influenza-like-illness levels from August 2012 to March 2019 for 47 prefectures in Japan collected from the Weekly Infectious Diseases Report in Japan.
    \item US-Regions: This dataset includes weekly influenza patient counts from 2002 to 2017 for 10 Health and Human Services regions in the United States collected from the Department of Health and Human Services.
    \item US-States: This dataset contains weekly influenza-positive cases from 2010 to 2017 for 49 states in the United States from the Center for Disease Control.
    \item Australia-COVID: This dataset obtained from the publicly accessible JHU-CSSE repository comprises daily counts of newly confirmed COVID-19 cases in Australia, encompassing all six states and two territories, spanning the period from January 27, 2020, to August 4, 2021.
\end{itemize}
The statistics of these four datasets are shown in Table \ref{tab:dataset_stats}.
We use Root Mean Squared Error (RMSE) and PCC to measure the difference between predicted and ground truth values:
\begin{equation}
\begin{gathered}
\mathrm{RMSE}=\sqrt{\frac{1}{N} \sum_{i=1}^N\left(\hat{y}_i-y_i\right)^2}, \\
\mathrm{PCC}=\frac{\sum_{i=1}^N\left(\hat{y}_i-\overline{\hat{y}}\right)\left(y_i-\bar{y}\right)}{\sqrt{\sum_{i=1}^N\left(\hat{y}_i-\overline{\hat{y}}\right)^2} \sqrt{\sum_{i=1}^N\left(y_i-\bar{y}\right)^2}},
\end{gathered}
\end{equation}
where $\hat{y}$ is the predicted values, $y$ is the ground truth values, and $\overline{\hat{y}}$ and $\bar{y}$ are their average in $N$ samples respectively.

\begin{table}[h!]
\vspace{-1mm}
\centering
\caption{Statistics of the four real-world datasets.\label{tab:dataset_stats}}
\scalebox{0.8}{
\begin{tabular}{c|ccccc}
\hline
Dataset & Size & Min & Max & Mean & Standard Deviation \\
\hline
Japan-Prefectures & $47 \times 348$ & 0 & 26635 & 655 & 1711 \\
US-Regions & $10 \times 785$ & 0 & 16526 & 1009 & 1351 \\
US-States & $49 \times 360$ & 0 & 9716 & 223 & 428 \\
Australia-COVID & $8 \times 556$ & 0 & 9987 & 539 & 1532 \\
\hline
\end{tabular}
}
\vspace{-3mm}
\end{table}

\textbf{Comparison Methods}. We compare four types of models: The traditional models include SIR, AutoRegressive (AR), AutoRegressive Moving Average (ARMA), Generalized AutoRegressive (GAR), and Vector AutoRegression (VAR). The Recurrent Neural Network (RNN)-based and Convolutional Neural Network (CNN)-based models include RNN \cite{werbos1990backpropagation}, Long Short-Term Memory (LSTM) \cite{graves2012long}, Gated Recurrent Unit (GRU) \cite{dey2017gate}, RNN-Attn \cite{cheng2016long}, CNNRNN-Res \cite{wu2018deep}, and Long- and Short-term Time-series Network (LSTNet) \cite{lai2018modeling}. The STGNN models include STGCN \cite{yu2017spatio}, Multiresolution Graph Neural Networks (MGNN) \cite{hy2022temporal}, Temporal Multiresolution Graph Neural Networks (TMGNN) \cite{hy2022temporal}, Cross-location Attention-based Graph Neural Networks (Cola-GNN) \cite{deng2020cola}, and Epi-GNN \cite{xie2022epignn}. The epidemiological machine learning model is Epi-Cola-GNN \cite{liu2023epidemiology}.

\textbf{Implementation Details.} We divide the dataset into training, validation, and test sets in chronological order with a ratio of 60\%-20\%-20\% \cite{wu2018deep,liu2023epidemiology}. We normalize all data based on the training data. The validation set is used for hyperparameter tuning and to prevent overfitting. The test set is used to evaluate the performance of the final model using RMSE and PCC \cite{wu2018deep}. The smaller the RMSE, the better the performance while the larger the PCC, the better the performance. We used Adam \cite{kingma2014adam} as the optimizer with weight decay 5e-4 during training. The grid search is used to find the best hyperparameter combinations. After the grid search, we set the initial learning rate to 0.001, the early stopping strategy with the patience of 200 epochs, the batch size to 32, and the number of epochs to 1500. All the parameters are initialized using Xavier initialization in HeatGNN. For all models, we set the historical window size $w$ to 20, that is, to use the historical data of the previous 20 weeks as input features. The horizon of advance forecasting $h$ is 2, 5, 7, and 12, that is, to predict the infection number of the $h$-th week after time step $t$. All experimental results are the average of 5 randomized trials. The experiments are conducted on an Intel Xeon Silver 4214 Processor CPU with 128 GB of 2666 MHz RAM and 4 NVIDIA TITAN RTX 24GB GPUs. Our method implemented in Python 3.8 and PyTorch 1.13.1 trains on GPUs in 30-60 minutes for each forecasting task.

\subsection{Prior Experiment\label{sec:prior}}
\begin{figure}[!htbp]
\centering
\includegraphics[width=0.47\textwidth]{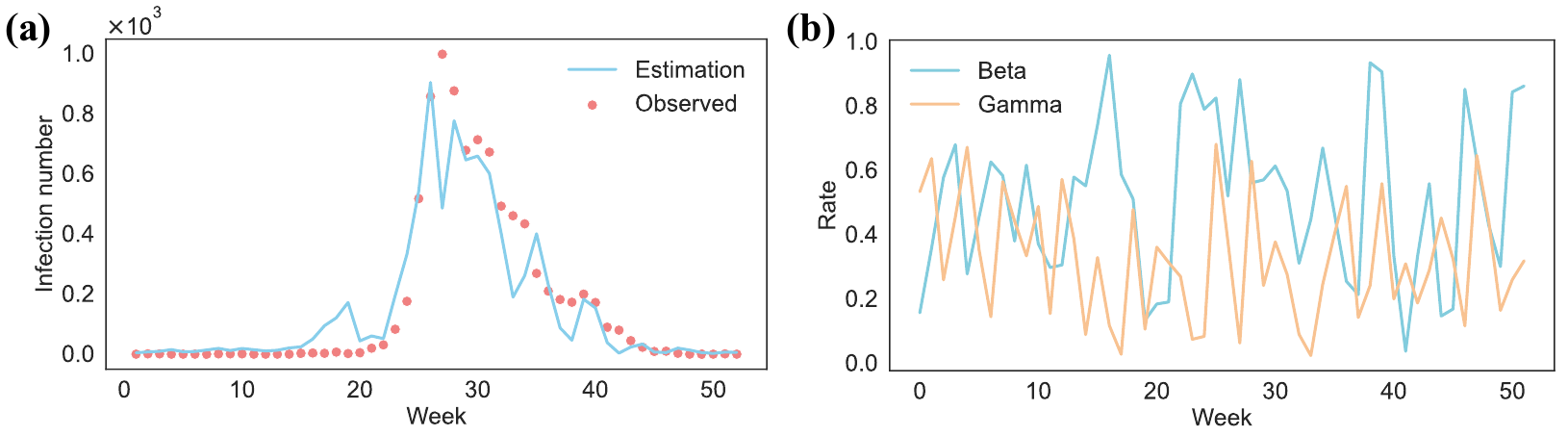}
\caption{Example of prior experiment. (a) is the estimation curve of infection number. (b) is the change of estimation parameter.\label{fig:prior_exp}}
\vspace{-6mm}
\end{figure}

To explore the heterogeneity of the spread of the epidemic in different regions, we design a prior experiment. We use the time-varying SIR model (Equation \ref{eq:t-sir}) to estimate the changes in the number of infections and the changes in its time-varying parameters in each location in Japan-Prefectures in the first 54 weeks (1 year). Based on previous studies, we use the simulated annealing algorithm to estimate the time-varying parameters (infection rate $\beta$ and recovery rate $\gamma$) in the time-varying SIR model. Note that each location is modeled and estimated separately. The experimental results show that the PCC between the true value and the estimated value of each location is basically above 0.9 (Table \ref{tab:stat_prior}). These results show that the fitting results are accurate and effective (Figure \ref{fig:prior_exp}(a)). After ensuring that the estimation results are reliable, we observe the heterogeneity of the time-varying parameters between different locations. We found that even if two locations have similar spatio-temporal transmission trends (e.g., infection number), there are differences in the trends of their internal mechanism transmission (e.g., infection rate and recovery rate). Some changes in the internal mechanism transmission are earlier than the easily observable change in spatio-temporal transmission (Figure \ref{fig:prior}). For example, Figure \ref{fig:prior_exp} shows the change in the infection number at one location and the change in the time-varying parameters (infection rate $\beta$ and recovery rate $\gamma$) estimated based on the time-varying SIR model. From week 10 to week 15, the estimated infection rate increased before the infection number increased, accompanied by a decrease in the recovery rate. 

Moreover, estimating time-varying parameters for all locations is computationally intensive. At each location, the two parameters to be estimated for the time-varying SIR model at a given moment are the infection rate and the recovery rate. If the parameters are estimated for all locations, with a time period $T$ and $N$ locations, the total number of estimated parameters is $2 \cdot T \cdot N$. Typically, the estimation of epidemic model parameters relies on metaheuristic algorithms, which present a significant challenge due to their computational complexity \cite{zhan2021identifying,zelenkov2023analysis}. This shows that it is unrealistic to use metaheuristic algorithms in deep learning models to estimate parameters for epidemic forecasting.

\begin{table}[!ht]
\vspace{-1mm}
\centering
\caption{Statistics of the fitting of the prior experiment.\label{tab:stat_prior}}
\begin{tabular}{c|cccc}
\hline
     & Mean   & Max     & Min    & Standard Deviation \\
\hline
PCC   & 0.9267 & 0.9983  & 0.8145 & 0.0462             \\
RMSE & 704.33 & 3400.30 & 18.039 & 825.43  \\
\hline
\end{tabular}
\vspace{-6mm}
\end{table}

\subsection{Overall Comparison (RQ1)}
\begin{table*}[!t]
\centering
\caption{Performance results on four datasets with horizon = 2, 5, 7, 12. A boldface indicates the best result, and the second-best is underlined. RMSE values are scaled by $\times 10^3$ to save space.}
\label{tab:main_results}
\scalebox{0.72}{
\begin{tabular}{c|cccccccc|cccccccc}
\hline
\multirow{3}{*}{Method} & \multicolumn{8}{c|}{Japan-Prefectures} & \multicolumn{8}{c}{US-Region} \\
\cline{2-17}
& \multicolumn{4}{c|}{RMSE ($\downarrow$)} & \multicolumn{4}{c|}{PCC ($\uparrow$)} & \multicolumn{4}{c|}{RMSE ($\downarrow$)} & \multicolumn{4}{c}{PCC ($\uparrow$)} \\
\cline{2-17}
& 2 & 5 & 7 & \multicolumn{1}{c|}{12} & 2 & 5 & 7 & 12 & 2 & 5 & 7 & \multicolumn{1}{c|}{12} & 2 & 5 & 7 & 12 \\
\hline
SIR                      & 78.173          & 217.438          & 432.737          & \multicolumn{1}{c|}{1991.195}          & 0.071          & -0.077          & -0.124          & -0.150          & 71.304          & 125.453          & 182.120          & \multicolumn{1}{c|}{446.533}          & 0.533          & 0.422          & 0.348          & 0.183          \\
AR                      & 1.670          & 2.440          & 2.561          & \multicolumn{1}{c|}{2.485}          & 0.743          & 0.303          & 0.154          & 0.491          & 0.626          & 1.079          & 1.280          & \multicolumn{1}{c|}{1.496}          & 0.921          & 0.788          & 0.705          & 0.583          \\
ARMA                    & 1.661          & 2.439          & 2.559          & \multicolumn{1}{c|}{2.482}          & 0.745          & 0.303          & 0.154          & 0.495          & 0.622          & 1.058          & 1.263          & \multicolumn{1}{c|}{1.484}          & 0.920          & 0.786          & 0.699          & 0.576          \\
GAR                     & 1.484          & 2.428          & 2.561          & \multicolumn{1}{c|}{2.530}          & 0.803          & 0.318          & 0.152          & 0.413          & 0.607          & 1.054          & 1.288          & \multicolumn{1}{c|}{1.568}          & 0.924          & 0.782          & 0.690          & 0.520          \\
VAR                     & 1.577          & 2.566          & 2.530          & \multicolumn{1}{c|}{2.504}          & 0.771          & 0.229          & 0.170          & 0.359          & 0.695          & 1.113          & 1.234          & \multicolumn{1}{c|}{1.380}           & 0.895          & 0.715          & 0.636          & 0.529          \\
RNN                     & 1.198          & 1.852          & 2.193          & \multicolumn{1}{c|}{1.906}          & 0.887          & 0.757          & 0.593          & 0.726          & 0.572          & 0.980          & 1.154          & \multicolumn{1}{c|}{1.513}          & 0.934          & 0.819          & 0.756          & 0.537          \\
LSTM                    & 1.255          & 1.840          & 2.235          & \multicolumn{1}{c|}{1.952}          & 0.884          & 0.820          & 0.606          & 0.723          & 0.570          & 1.049          & 1.230          & \multicolumn{1}{c|}{1.531}          & 0.933          & 0.820           & 0.754          & 0.580          \\
GRU                     & 1.181          & 1.782          & 2.226          & \multicolumn{1}{c|}{1.922}          & 0.894          & 0.807          & 0.599          & 0.725          & 0.571          & 1.024          & 1.205          & \multicolumn{1}{c|}{1.521}          & 0.933          & 0.817          & 0.753          & 0.527          \\
RNN-Attn                & 1.311          & 2.267          & 2.133          & \multicolumn{1}{c|}{1.933}          & 0.865          & 0.497          & 0.662          & 0.715          & 0.566          & 0.999          & 1.211          & \multicolumn{1}{c|}{1.521}          & 0.936          & 0.820          & 0.734          & 0.567          \\
CNNRNN-Res              & 1.401          & 2.570          & 2.603          & \multicolumn{1}{c|}{2.067}          & 0.828          & 0.239          & 0.076          & 0.654          & 0.595          & 1.006          & 1.153          & \multicolumn{1}{c|}{1.316}          & 0.925          & 0.777          & 0.687          & 0.600          \\
LSTNet                  & 1.441          & 2.385          & 2.477          & \multicolumn{1}{c|}{2.391}          & 0.820          & 0.338          & 0.267          & 0.438          & 0.607          & 1.134          & 1.121          & \multicolumn{1}{c|}{1.149}          & 0.934          & 0.733          & 0.748          & 0.682          \\
STGCN                   & 1.274          & \underline{1.400}    & \textbf{1.478} & \multicolumn{1}{c|}{1.812}          & 0.892          & \underline{0.872}    & \textbf{0.853} & 0.761          & 0.727          & 0.965          & 1.017          & \multicolumn{1}{c|}{1.116}          & 0.899          & 0.831          & 0.817          & 0.733          \\
MGNN                    & 1.757          & 1.943          & 2.048          & \multicolumn{1}{c|}{2.655}          & 0.688          & 0.507          & 0.400          & 0.133          & 1.472          & 1.487          & 1.305          & \multicolumn{1}{c|}{1.538}          & 0.558          & 0.550          & 0.549          & 0.536          \\
TMGNN                   & 1.500          & 1.559          & 1.977          & \multicolumn{1}{c|}{1.839}          & 0.827          & 0.822          & 0.721          & 0.777          & 0.656          & 0.916          & 1.043          & \multicolumn{1}{c|}{1.228}          & 0.913          & 0.833          & 0.808          & 0.796          \\
Cola-GNN                & 1.168          & 1.573          & 1.758          & \multicolumn{1}{c|}{\underline{1.690}}    & 0.906          & 0.834          & \underline{0.787}    & \textbf{0.804} & 0.552          & \underline{0.871}    & 1.054          & \multicolumn{1}{c|}{\textbf{1.011}} & 0.939          & 0.847          & 0.805          & 0.780          \\
Epi-Cola-GNN            & \textbf{1.118} & 1.544          & 2.549          & \multicolumn{1}{c|}{2.164}          & \underline{0.912}    & 0.859          & 0.179          & 0.679          & \underline{0.547}    & 0.872          & 1.160          & \multicolumn{1}{c|}{1.365}          & \underline{0.940}    & \underline{0.858}    & 0.751          & 0.648          \\
Epi-GNN                 & 1.503          & 1.448          & 1.738          & \multicolumn{1}{c|}{1.695}          & 0.821          & 0.869          & 0.768          & \underline{0.786}    & 0.563          & 0.881          & \underline{0.938}    & \multicolumn{1}{c|}{1.025}          & 0.937          & \underline{0.858}    & \underline{0.849}    & \underline{0.821}    \\
HeatGNN                 & \underline{1.149}    & \textbf{1.378} & \underline{1.735}    & \multicolumn{1}{c|}{\textbf{1.685}} & \textbf{0.917} & \textbf{0.884} & 0.780          & 0.773          & \textbf{0.541} & \textbf{0.852} & \textbf{0.922} & \multicolumn{1}{c|}{\underline{1.024}}    & \textbf{0.941} & \textbf{0.866} & \textbf{0.856} & \textbf{0.824} \\ \hline
\multirow{3}{*}{Method} & \multicolumn{8}{c|}{US-State} & \multicolumn{8}{c}{Australia-COVID} \\
\cline{2-17}
& \multicolumn{4}{c|}{RMSE ($\downarrow$)} & \multicolumn{4}{c|}{PCC ($\uparrow$)} & \multicolumn{4}{c|}{RMSE ($\downarrow$)} & \multicolumn{4}{c}{PCC ($\uparrow$)} \\
\cline{2-17}
& 2 & 5 & 7 & \multicolumn{1}{c|}{12} & 2 & 5 & 7 & 12 & 2 & 5 & 7 & \multicolumn{1}{c|}{12} & 2 & 5 & 7 & 12 \\
\hline
SIR                       & 17.872          & 31.446          & 45.436          & \multicolumn{1}{c|}{110.536}          & 0.656          & 0.584          & 0.537          & 0.431          & 74.009          & 112.709          & 144.578          & \multicolumn{1}{c|}{231.556}                & 0.996          & 0.995          & 0.993          & 0.989          \\
AR                       & 0.172          & 0.264          & 0.300          & \multicolumn{1}{c|}{0.341}          & 0.934          & 0.850          & 0.795          & 0.717          & 0.509          & 0.540          & 0.556          & \multicolumn{1}{c|}{0.587}                & 0.990          & 0.989          & 0.988          & 0.986          \\
ARMA                     & 0.172          & 0.262          & 0.298          & \multicolumn{1}{c|}{0.340}          & 0.933          & 0.849          & 0.796          & 0.718          & 0.505          & 0.527          & 0.541          & \multicolumn{1}{c|}{0.564}                & 0.989          & 0.988          & 0.987          & 0.986          \\
GAR                      & 0.162          & 0.247          & 0.292          & \multicolumn{1}{c|}{0.355}          & 0.938          & 0.861          & 0.805          & 0.708          & 1.343          & 1.394          & 1.430          & \multicolumn{1}{c|}{1.519}                & 0.733          & 0.727          & 0.723          & 0.700          \\
VAR                      & 0.298          & 0.333          & 0.397          & \multicolumn{1}{c|}{0.422}          & 0.779          & 0.730          & 0.685          & 0.581          & 0.581          & 0.592          & 0.607          & \multicolumn{1}{c|}{0.619}                & 0.984          & 0.982          & 0.981          & 0.978          \\
RNN                      & 0.159          & 0.224          & 0.255          & \multicolumn{1}{c|}{0.329}          & 0.944          & 0.890          & 0.861          & 0.747          & 0.581          & 0.648          & 0.708          & \multicolumn{1}{c|}{0.898}                & 0.992          & 0.991          & 0.991          & \underline{0.990}     \\
LSTM                     & 0.161          & 0.227          & 0.263          & \multicolumn{1}{c|}{0.330}          & 0.943          & 0.892          & 0.853          & 0.749          & 1.122          & 1.178          & 1.230          & \multicolumn{1}{c|}{1.393}                & 0.985          & 0.984          & 0.984          & 0.982          \\
GRU                      & 0.159          & 0.248          & 0.285          & \multicolumn{1}{c|}{0.335}          & 0.944          & 0.870          & 0.822          & 0.742          & 1.065          & 1.127          & 1.177          & \multicolumn{1}{c|}{1.322}                & \underline{0.995}    & \underline{0.994}    & \textbf{0.993} & \textbf{0.991} \\
RNN-Attn                 & 0.163          & 0.244          & 0.274          & \multicolumn{1}{c|}{0.342}          & 0.943          & 0.869          & 0.823          & 0.711          & 1.054          & 1.039          & 1.026          & \multicolumn{1}{c|}{1.166}                & 0.862          & 0.867          & 0.868          & 0.717          \\
CNNRNN-Res               & 0.228          & 0.320          & 0.360          & \multicolumn{1}{c|}{0.280}          & 0.887          & 0.731          & 0.649          & 0.808          & 1.484          & 1.488          & 1.491          & \multicolumn{1}{c|}{1.476}                & 0.959          & 0.962          & 0.961          & 0.964          \\
LSTNet                   & 0.196          & 0.276          & 0.288          & \multicolumn{1}{c|}{0.312}          & 0.924          & 0.811          & 0.785          & 0.794          & 0.595          & 0.614          & 0.645          & \multicolumn{1}{c|}{0.712}                & 0.990          & 0.989          & 0.988          & 0.984          \\
STGCN                    & 0.205          & 0.260          & 0.280          & \multicolumn{1}{c|}{0.296}          & 0.901          & 0.838          & 0.808          & 0.804          & 0.696          & 0.649          & 0.706          & \multicolumn{1}{c|}{0.721}                & 0.960          & 0.961          & 0.964          & 0.971          \\
MGNN                     & 0.178          & 0.206          & 0.259          & \multicolumn{1}{c|}{0.288}    & 0.926          & 0.901          & 0.870          & 0.851          & 0.664          & 0.769          & 0.848          & \multicolumn{1}{c|}{0.857}                & 0.826          & 0.823          & 0.820          & 0.818          \\
TMGNN                    & 0.179          & \underline{0.190}    & 0.438    & \multicolumn{1}{c|}{0.447} & 0.925          & 0.913          & 0.549          & 0.545 & 0.803          & 0.846          & 0.870          & \multicolumn{1}{c|}{0.916}                & 0.829          & 0.828          & 0.827          & 0.824          \\
Cola-GNN                 & \underline{0.148}    & 0.224          & 0.228          & \multicolumn{1}{c|}{\underline{0.241}}          & \underline{0.953}    & 0.904          & 0.898          & \textbf{0.893}    & 0.368          & 0.406          & 0.463          & \multicolumn{1}{c|}{0.533}                & 0.993          & \textbf{0.994} & 0.990          & 0.988          \\
Epi-Cola-GNN             & \underline{0.148}    & 0.219          & 0.268          & \multicolumn{1}{c|}{0.303}          & \underline{0.953}    & 0.909          & 0.850          & 0.819          & 0.803    & 0.854    & 0.912 & \multicolumn{1}{c|}{1.124}          & 0.974    & 0.975          & 0.957          & 0.970          \\
Epi-GNN                  & 0.150          & \underline{0.190}    & \underline{0.205}    & \multicolumn{1}{c|}{0.251}          & 0.952          & \textbf{0.923} & \underline{0.909}    & \underline{0.881}          & \underline{0.325}          & \underline{0.381}          & \underline{0.416}          & \multicolumn{1}{c|}{\underline{0.491}}          & \textbf{0.995} & 0.993          & \underline{0.992}    & 0.989          \\
HeatGNN                  & \textbf{0.142} & \textbf{0.186} & \textbf{0.200} & \multicolumn{1}{c|}{\textbf{0.240}}          & \textbf{0.953} & \underline{0.921}    & \textbf{0.911} & 0.870          & \textbf{0.315} & \textbf{0.334} & \textbf{0.411}    & \multicolumn{1}{c|}{\textbf{0.466}} & 0.993          & 0.993          & 0.991          & 0.986          \\
\hline
\end{tabular}
}
\vspace{-3mm}
\end{table*}

\begin{figure}[!htbp]
\centering
\includegraphics[width=0.4\textwidth]{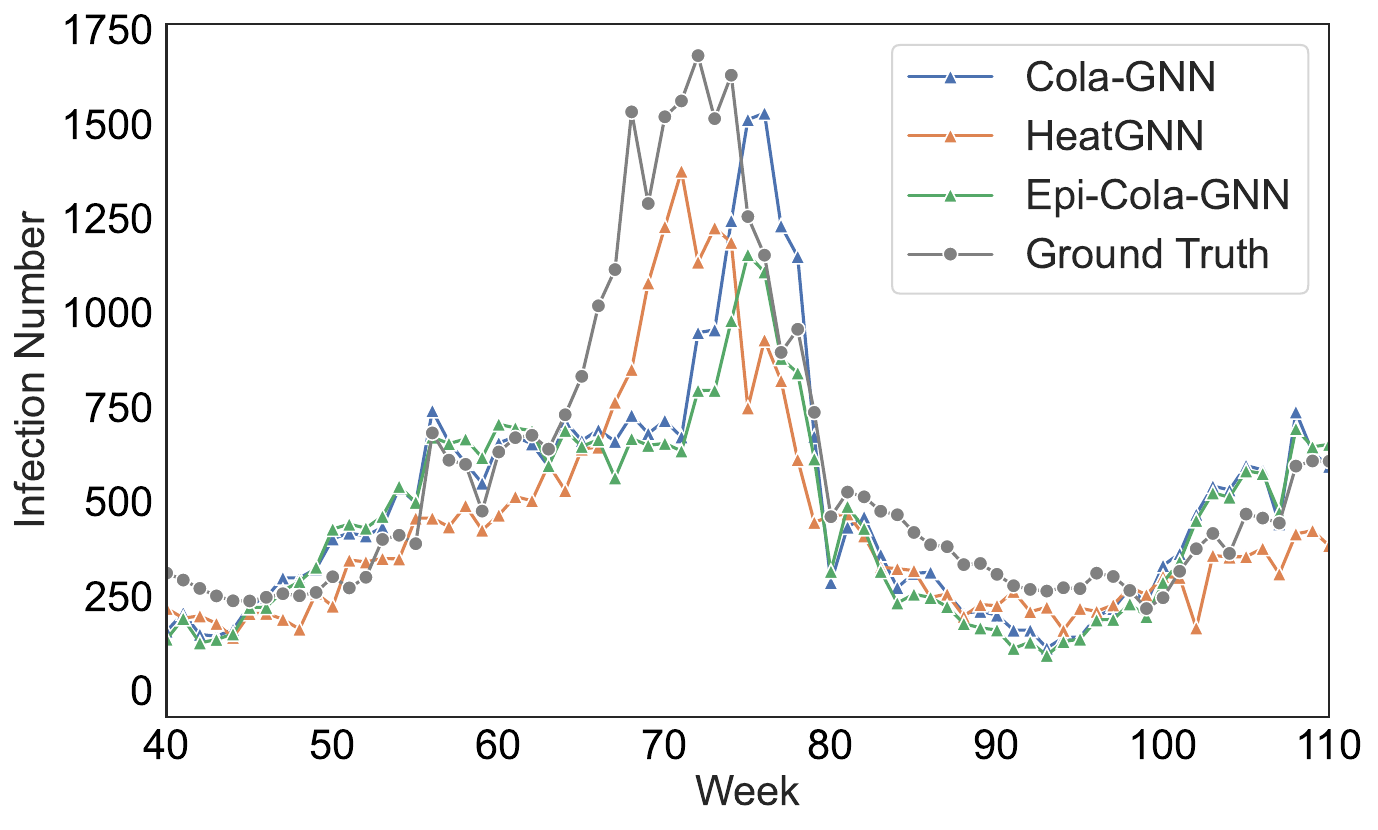}
\caption{Visualization of prediction and ground truth values under horizon=5 in US-Regions.\label{fig:case_region5}}
\vspace{-6mm}
\end{figure}

We evaluate our method in both short-term forecasting (horizon=2, 5) and long-term forecasting (horizon=7, 12) settings. Table \ref{tab:main_results} summarizes the results of all methods under four real-world datasets, where we scale the RMSE to save space. Compared with various baseline models, including traditional, RNN-based, CNN-based, STGNN, and EML models, HeatGNN achieves the lowest RMSE and highest PCC across most horizons and datasets. These results demonstrate its capability to capture both spatio-temporal heterogeneity and mechanistic heterogeneity effectively. The high PCC values across datasets and horizons highlight the robustness of HeatGNN in capturing transmission patterns that are closely related to real-world data, making it well-suited for short-term and long-term epidemic forecasting. Among the other models, Epi-Cola-GNN and Epi-GNN show the second-best performances. Figure \ref{fig:case_region5} shows that HeatGNN more accurately fits the ground truth and follows the epidemic transmission trend.

As the horizon increases, HeatGNN maintains competitive performance, while several models show significant error growth. Short-term forecasting performs well due to strong temporal dependencies, as evidenced by good forecasting performance for all models at horizon=2. However, traditional and CNN-based models lose temporal dependency information as the horizon extends, leading to performance declines. In contrast, the RNN-based models maintain relatively robust performance with their memory of temporal dependencies. The improved performance of STGNN models, such as Cola-GNN, STGCN, and TMGNN, highlights the importance of spatio-temporal heterogeneity in forecasting. However, while spatio-temporal dependencies enhance the ability to capture dynamic changes in epidemic transmission, they also introduce added complexity, compromising the robustness of specific models, such as MGNN and TMGNN. Additionally, incorporating epidemic mechanism models into GNNs improves short-term forecasting accuracy. For instance, Epi-Cola-GNN achieves the second-best performance across three datasets at horizon=2.

\subsection{Ablation Study (RQ2)}
We conduct the ablation study by testing the following variants of HeatGNN: (1) “w/o PL”, remove the physics loss from HeatGNN, i.g. remove the constraints on the EIEL module; (2) “w/o TG”, remove the TG module and use only the STGL module for forecasting while retaining the EIEL module to guide the output of the framework; (3) “w/o TG+EIEL”, remove the EIEL and TG modules and only keep the STGL module for forecasting due to the TG module being based on the EIEL module.

The ablation study results of RMSE and PCC are shown in Table \ref{tab:ablation}. The results show that our proposed HeatGNN performs robustly across most cases. The physical loss incorporates domain knowledge of epidemic transmission, particularly benefiting high-variability regions (e.g., Japan-Prefectures and US-Regions) and long-term forecasting. Removing the physics loss (“w/o PL”) results in the model lacking intrinsic epidemic transmission mechanism constraints, which increases the forecasting error and reduces its ability to capture trends. However, its impact is limited in the US-States dataset, which has low volatility and relatively uniform data. In fact, the “w/o PL” variant performs comparably to HeatGNN on the US-State dataset at some forecast horizons, suggesting that the benefits of physical loss's regularization are less significant in more stable, homogeneous regions.

The TG module is crucial for modeling cross-regional epidemic transmission. After removing the TG module (“w/o TG”), the model is significantly weaker at capturing the intrinsic epidemic transmission relationship between locations. While it has a similar role to the physics loss across datasets, its impact on performance is more pronounced. This is particularly evident in the US-Region and US-State datasets, where the “w/o TG” leads to a significant drop in PCC and an increase in RMSE. Moreover, on the Australia-COVID dataset, “w/o TG” slightly improves RMSE at longer horizons, suggesting that TG may introduce redundancy when inter-regional transmission is weak, while PCC still drops, indicating that TG remains useful for capturing trend consistency.

Furthermore, the EIEL module enhances the effectiveness of the TG module, further improving the ability to capture epidemic transmission. Relying solely on the STGL module is insufficient for addressing complex epidemic forecasting tasks. When both TG and EIEL modules are removed (“w/o TG+EIEL”), the model struggles to accurately capture epidemic transmission over extended time horizons, resulting in a significant decline in forecasting performance. These findings highlight the complementary and synergistic roles of each module. The complete HeatGNN consistently achieves the best result in most cases and forecast horizons, demonstrating the importance of each component and its integration.

\begin{table*}[thbp]
\centering
\caption{Ablation test results on four datasets: Japan-Prefectures, US-Region, US-State, and Australia-COVID.\label{tab:ablation}}
\scalebox{0.75}{
\begin{tabular}{c|cccccccc|cccccccc}
\hline
\multirow{3}{*}{Method} & \multicolumn{8}{c|}{Japan-Prefectures}                                                                                                                    & \multicolumn{8}{c}{US-Region}                                                                                                                              \\ \cline{2-17} 
                        & \multicolumn{4}{c|}{RMSE ($\downarrow$)}                                                              & \multicolumn{4}{c|}{PCC ($\uparrow$)}                                         & \multicolumn{4}{c|}{RMSE ($\downarrow$)}                                                              & \multicolumn{4}{c}{PCC ($\uparrow$)}                                           \\ \cline{2-17} 
                        & 2              & 5              & 7              & \multicolumn{1}{c|}{12}             & 2              & 5              & 7             & 12             & 2              & 5              & 7              & \multicolumn{1}{c|}{12}             & 2              & 5              & 7              & 12             \\ \hline
w/o PL                  & 1.230          & 1.453          & 1.804          & \multicolumn{1}{c|}{1.759}          & 0.900          & 0.863          & 0.746         & 0.757          & 0.631          & 0.943          & 1.055          & \multicolumn{1}{c|}{1.141}          & 0.924          & 0.864          & 0.840          & 0.823          \\
w/o TG                  & 1.229          & 1.439          & 1.762          & \multicolumn{1}{c|}{1.657}          & 0.891          & 0.871          & 0.783         & 0.754          & 0.740          & 1.039          & 1.142          & \multicolumn{1}{c|}{1.182}          & 0.923          & 0.838          & 0.797          & 0.780          \\
w/o TG+EIEL             & 1.279          & 1.452          & 1.765          & \multicolumn{1}{c|}{1.675}          & 0.890          & 0.853          & 0.774         & 0.760          & 0.727          & 0.977          & 1.119          & \multicolumn{1}{c|}{1.188}          & 0.927          & 0.854          & 0.818          & 0.790          \\
HeatGNN                 & \textbf{1.149} & \textbf{1.378} & \textbf{1.735} & \multicolumn{1}{c|}{\textbf{1.685}} & \textbf{0.917} & \textbf{0.884} & \textbf{0.780} & \textbf{0.773} & \textbf{0.541} & \textbf{0.852} & \textbf{0.922} & \multicolumn{1}{c|}{\textbf{1.024}} & \textbf{0.941} & \textbf{0.866} & \textbf{0.856} & \textbf{0.824} \\
\hline
\multirow{3}{*}{Method} & \multicolumn{8}{c|}{US-State}                                                                                                                              & \multicolumn{8}{c}{Australia-COVID}                                                                                                                        \\ \cline{2-17} 
                        & \multicolumn{4}{c|}{RMSE ($\downarrow$)}                                                              & \multicolumn{4}{c|}{PCC ($\uparrow$)}                                          & \multicolumn{4}{c|}{RMSE ($\downarrow$)}                                                              & \multicolumn{4}{c}{PCC ($\uparrow$)}                                           \\ \cline{2-17} 
                        & 2              & 5              & 7              & \multicolumn{1}{c|}{12}             & 2              & 5              & 7              & 12             & 2              & 5              & 7              & \multicolumn{1}{c|}{12}             & 2              & 5              & 7              & 12             \\ \hline
w/o PL                  & 0.146          & 0.188          & \textbf{0.192} & \multicolumn{1}{c|}{\textbf{0.223}} & 0.951          & 0.917          & \textbf{0.912} & \textbf{0.879} & 0.487          & 0.460          & 0.451          & \multicolumn{1}{c|}{0.455}          & 0.986          & 0.985          & 0.985          & 0.987          \\
w/o TG                  & 0.149          & 0.202          & 0.215          & \multicolumn{1}{c|}{0.238}          & 0.951          & 0.907          & 0.891          & 0.858          & 0.350          & 0.366          & \textbf{0.362} & \multicolumn{1}{c|}{\textbf{0.429}} & 0.991          & 0.991          & 0.991          & 0.986          \\
w/o TG+EIEL             & 0.153          & 0.206          & 0.217          & \multicolumn{1}{c|}{0.227}          & 0.951          & 0.902          & 0.892          & 0.880          & 0.461          & 0.481          & 0.449          & \multicolumn{1}{c|}{0.435}          & 0.984          & 0.986          & 0.988          & \textbf{0.987} \\
HeatGNN                 & \textbf{0.142} & \textbf{0.186} & 0.200          & \multicolumn{1}{c|}{0.240}          & \textbf{0.953} & \textbf{0.921} & 0.911          & 0.870          & \textbf{0.315} & \textbf{0.334} & 0.411          & \multicolumn{1}{c|}{0.466}          & \textbf{0.993} & \textbf{0.993} & \textbf{0.991} & 0.986          \\
\hline
\end{tabular}
}
\vspace{-6mm}
\end{table*}

\begin{figure*}[!htbp]
\centering
\includegraphics[width=0.8\textwidth]{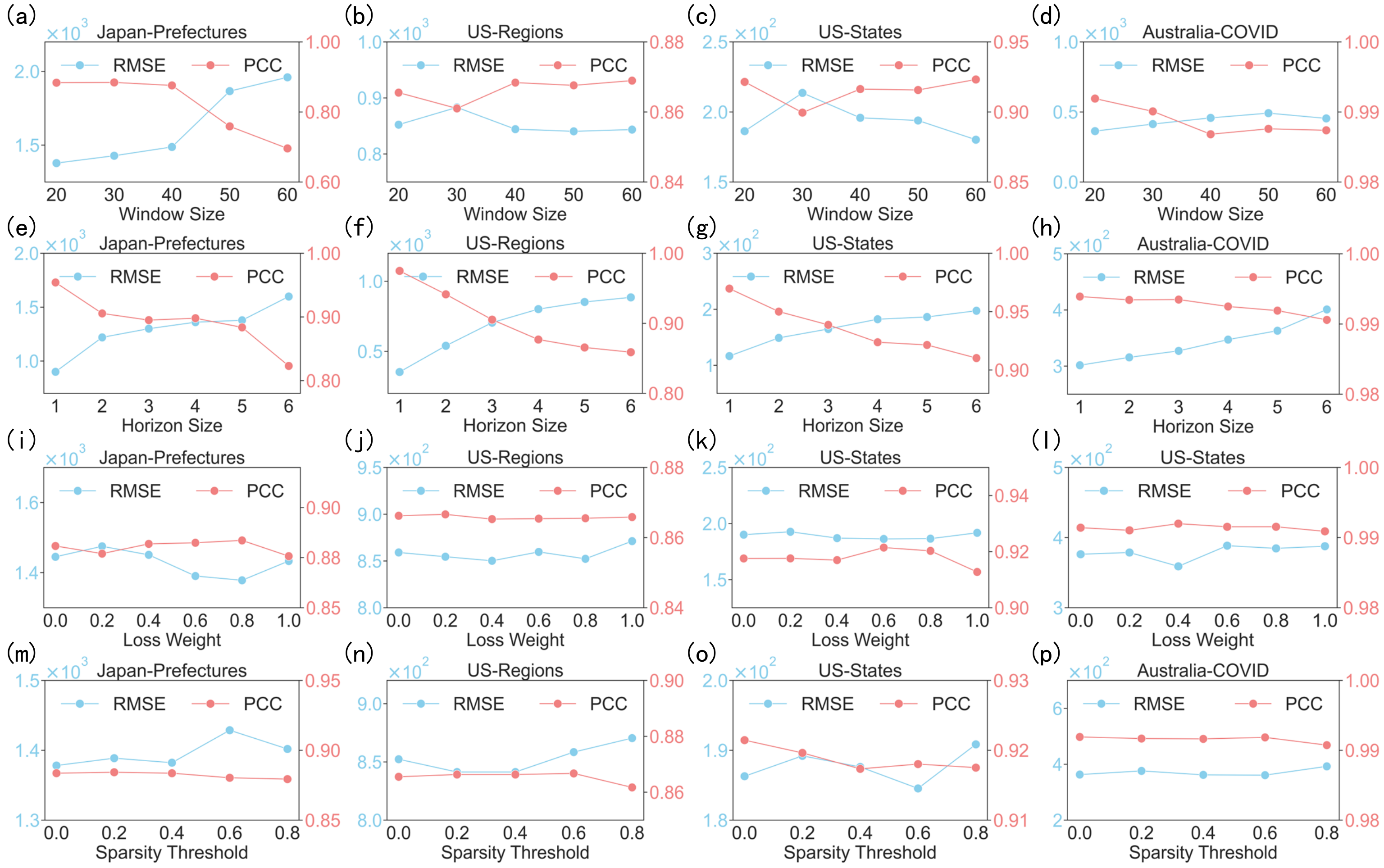}
\caption{Sensitivity analysis on hyperparameter. Performance with window size $w$, horizon size $h$, loss weight $\lambda$, and sparsity threshold $\delta$. Except for the sensitivity analysis of $h$, other sensitivity analyses are conducted under horizon=5.\label{fig:sen}}
\vspace{-6mm}
\end{figure*}

\subsection{Hyperparameter Sensitivity Analysis (RQ3)}
We test how performance varies with hyperparameters.

\textbf{Window size $w$.}
To evaluate whether the model is sensitive to the length of historical data, we evaluate different window sizes ranging from 20 to 60. Figures \ref{fig:sen} (a), (b), (c), and (d) show the RMSE and PCC results in different window sizes. For the Japan-Prefectures and Australia-COVID datasets, forecasting performance improves significantly with larger window sizes, as evidenced by a decrease in RMSE and an increase in PCC. In contrast, model performance remains stable across all window sizes for the US-Regions dataset, which indicates that the model is not sensitive to the window size change in this dataset. For the US-States dataset, the optimal performance is observed at a window size of 60. In summary, while increasing the window size enhances model performance, the improvement becomes limited after a threshold.

\textbf{Horizon size $h$.}
To evaluate the sensitivity of the model to the size of the forecasting horizon, we analyze its performance across horizons ranging from 1 to 6. Figures \ref{fig:sen} (e), (f), (g), and (h) show the RMSE and PCC results for different horizons. In the Japan-Prefectures dataset, as the horizon increases, RMSE gradually rises while PCC decreases. Similar trends are observed in the US-Regions, US-States, and Australia-COVID datasets. Comparing the results across the four datasets, we find that datasets with more significant fluctuations (Japan-Prefectures), shorter horizons, and more accurate predictions, while datasets with lower variability (US-Regions, US-States, and Australia-COVID) maintain robust forecasting performance even with longer horizons.

\textbf{Loss weight $\lambda$.}
To evaluate the sensitivity of the model to the balance weight between physical and data loss, we examine the effect of varying $\lambda$ values from 0 to 1 on forecasting performance. Figures \ref{fig:sen} (i), (j), (k), and (l) show the RMSE and PCC results under different $\lambda$. Overall, increasing the $\lambda$ value has a greater impact on datasets with greater variability, while datasets with lower variability (e.g., US-States) maintain good performance even at higher $\lambda$ values. The results suggest that reasonably adjusting the balance between physical and data loss can enhance model performance across different datasets.

\textbf{Sparsity Threshold $\delta$.}
To evaluate whether the model is sensitive to the sparsification of the MAG, we test the impact of different sparsity thresholds, from 0 to 0.8 (Figures \ref{fig:sen} (m), (n), (o), and (p)). A higher threshold corresponds to a sparser MAG. On the Japan-Prefectures dataset, as the sparsity threshold increases, RMSE fluctuates, indicating that excessive sparsification may degrade the forecasting performance. On the US-Regions dataset, RMSE shows minor fluctuation, while PCC remains stable. For the US-States dataset, the model achieves the best performance at a lower sparsity threshold, with a slight decrease in PCC as the threshold increases. For the Australia-COVID dataset, the performance of different sparsity thresholds basically remains unchanged. Overall, a modest increase in the sparsity threshold improves model performance, and selecting an appropriate threshold helps balance data complexity and model effectiveness.

\subsection{Interpretability Analysis (RQ4)}
\begin{figure}[!tbp]
\vspace{-4mm}
\centering
\includegraphics[width=0.47\textwidth]{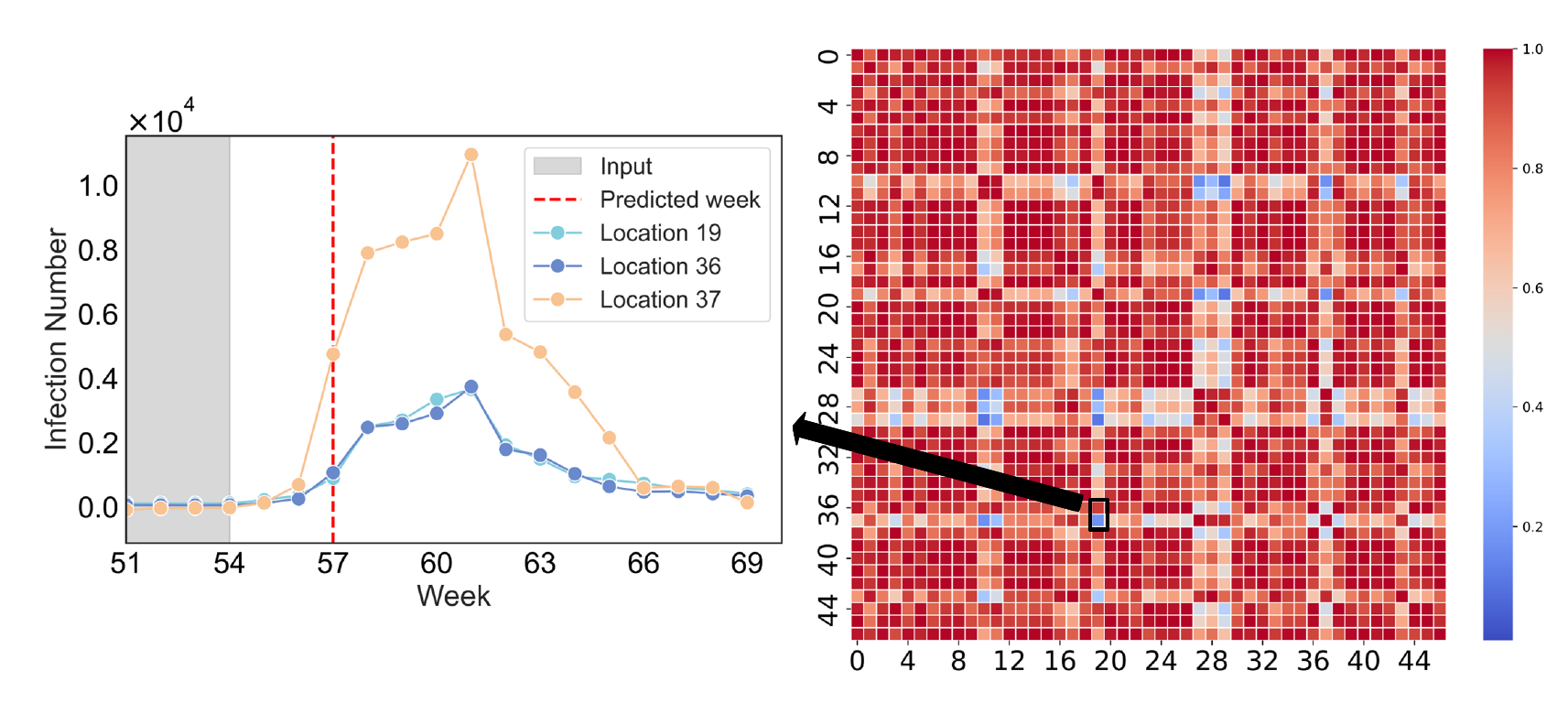}
\caption{Mechanistic similarity $\boldsymbol{M}_t$. On the left is the actual infection number change of three locations. On the right is the mechanism similarity matrix learned from historical data (up to week 54) for prediction in week 57.\label{fig:case}}
\vspace{-4mm}
\end{figure}

\begin{figure}[!tbp]
\centering
\includegraphics[width=0.45\textwidth]{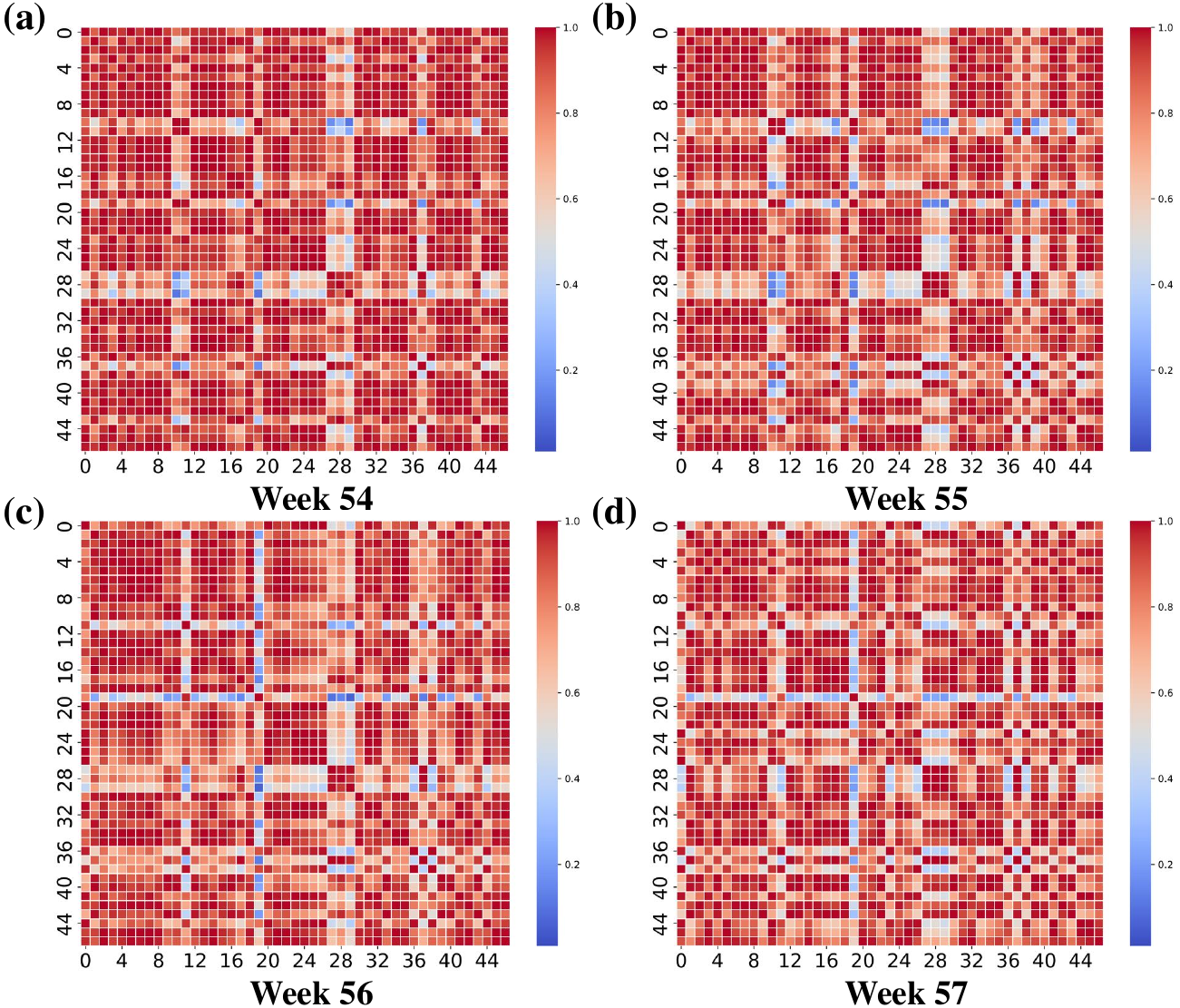}
\caption{Mechanistic similarity $\boldsymbol{M}_t$ in continuous timesteps.\label{fig:case_study_mag}}
\vspace{-6mm}
\end{figure}

We conduct a case study in the Japan-Prefectures dataset with horizon=3 and 47 nodes. Figures \ref{fig:case} and \ref{fig:case_study_mag} illustrate the mechanistic similarity $\boldsymbol{M}_{t}$ between different locations learned from the TG module in the Japan-Prefectures dataset at single and continuous timestep. In this analysis, we focus on week 54 and utilize historical data from week 54 and the previous 19 weeks (a total of 20 weeks) to predict the infection number three weeks later (week 57). By observing location 19 (prediction target), location 36, and location 37, the infection number trends of these three locations show similar patterns up to week 54. However, after week 54, the trend at location 37 diverges significantly, rising sharply in week 57 and reaching a higher peak in subsequent weeks. In contrast, locations 19 and 36 exhibit more similar trends, with a gradual increase in infection number that does not reach the same peak as location 37. The mechanism similarity, calculated using HeatGNN, highlights differences in future mechanism propagation across locations as of week 54. Specifically, the mechanism similarity reveals a lower similarity between location 19 and location 37 (indicated in blue) and a higher similarity between location 19 and location 36 (indicated in red). This finding demonstrates that the mechanism similarity effectively captures hidden dynamics between locations under the mechanism of spread.

\begin{figure}[!htbp]
\centering
\includegraphics[width=0.4\textwidth]{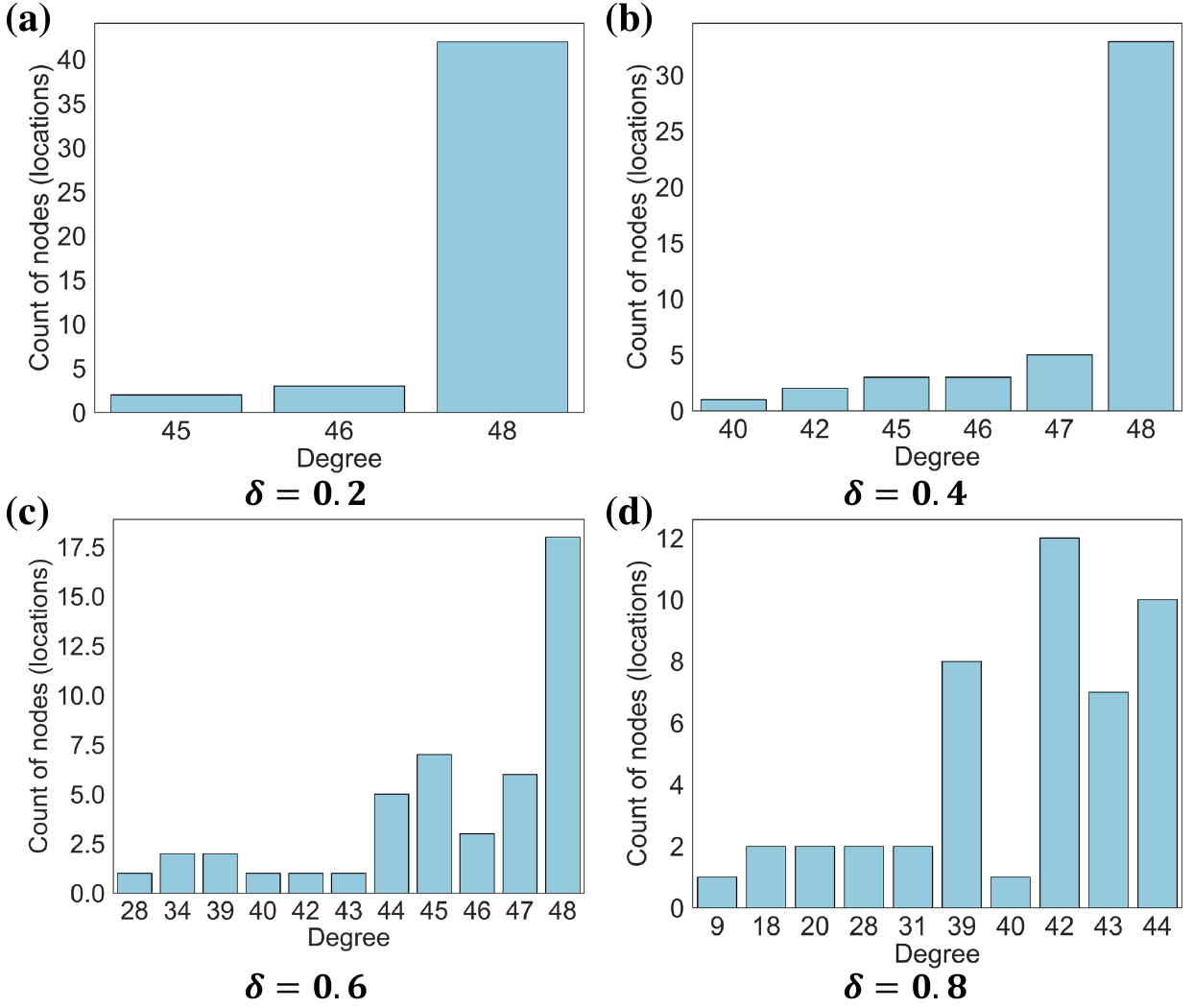}
\caption{The degree distribution at different thresholds in MAG.\label{fig:case_degree}}
\vspace{-4mm}
\end{figure}

We conduct a MAG comparison experiment to prove that the MAG used by the TG module better captures mechanistic heterogeneity: based on real data, we fit the SIR model for each location and construct a real MAG that reflects the heterogeneity of the real epidemic transmission mechanistic as a baseline (real MAG). Comparing this real MAG with the MAG generated by the EIEL module embedding ($MAG_{E}$) and the MAG generated by the ST module embedding ($MAG_{S}$) based on the Manhattan distance and the Euclidean distance, the results show that the MAG generated by the EIEL module is closer to the real MAG (Table \ref{tab:MAG}). This finding indicates that more correlation between EIEL module embedding and established epidemic factors. Furthermore, we calculate the average degree and degree distribution at different thresholds in MAG (Table \ref{tab:MAG_Degree} and Figure \ref{fig:case_degree}). The motivation behind MAG is to emphasize information propagation between mechanistically similar locations, which is enabled primarily by the similarity-weighted nature of edges. Meanwhile, the role of the threshold is mainly for efficient graph convolution by allowing MAG to be operated as a sparse matrix, thus having a lower impact on the performance. This is also verified by aligning Figure \ref{fig:sen} and Table \ref{tab:ablation}’s results - as long as MAG is in use (despite the value of $\delta$), the performance outperforms the “w/o TG” variant (MAG is completely disabled) in most cases.

\begin{table}[!b]
\vspace{-4mm}
    \centering
    \caption{Distances between the generated and real MAGs. \label{tab:MAG}}
    \begin{tabular}{c|cc}
    \hline
        ~ & $MAG_{E}$ & $MAG_{S}$ \\ \hline
        Manhattan distance $(\downarrow)$ & 9.94 & 7.29 \\
        Euclidean distance $(\downarrow)$ & 0.30 & 0.22 \\ \hline
    \end{tabular}
    \vspace{-2mm}
\end{table}

\begin{table}[!b]
    \centering
    \caption{Average degrees at different thresholds in MAG.\label{tab:MAG_Degree}}
    \begin{tabular}{c|ccccc}
    \hline
        Threshold $\delta$ & 0 & 0.2 & 0.4 & 0.6 & 0.8 \\ \hline
        Average Degree & 47.0 & 45.7 & 45.1 & 43.0 & 36.2 \\ \hline
    \end{tabular}
\end{table}


\subsection{Time Complexity and Scalability Analysis (RQ5)}
This section aims to test whether HeatGNN has robust scalability. We analyze the time complexity of the MAG in HeatGNN. Let $N$ be the number of positions and $N_f$ be the feature dimension of each position. The time complexity of MAG includes three parts: the complexity of feature normalization is $O(N \cdot N_f)$, which is used to standardize the features of each position; the calculation of the cosine similarity matrix in MAG requires similarity evaluation of all position pairs, with a complexity of $O(N^2 \cdot N_f)$; and the final MAG normalization operation is $O(N^2)$. In summary, the overall time complexity of MAG is $O(N^2 \cdot N_f)$, among which the cosine similarity calculation is dominant. We compare the inference time of HeatGNN with two baselines (Table \ref{tab:inference}). The result shows that HeatGNN performs well in inference efficiency, which is lower than other baselines in most cases, indicating that it has good practical application insight while maintaining low computational overhead.

\begin{figure}[!t]
\centering
\includegraphics[width=0.35\textwidth]{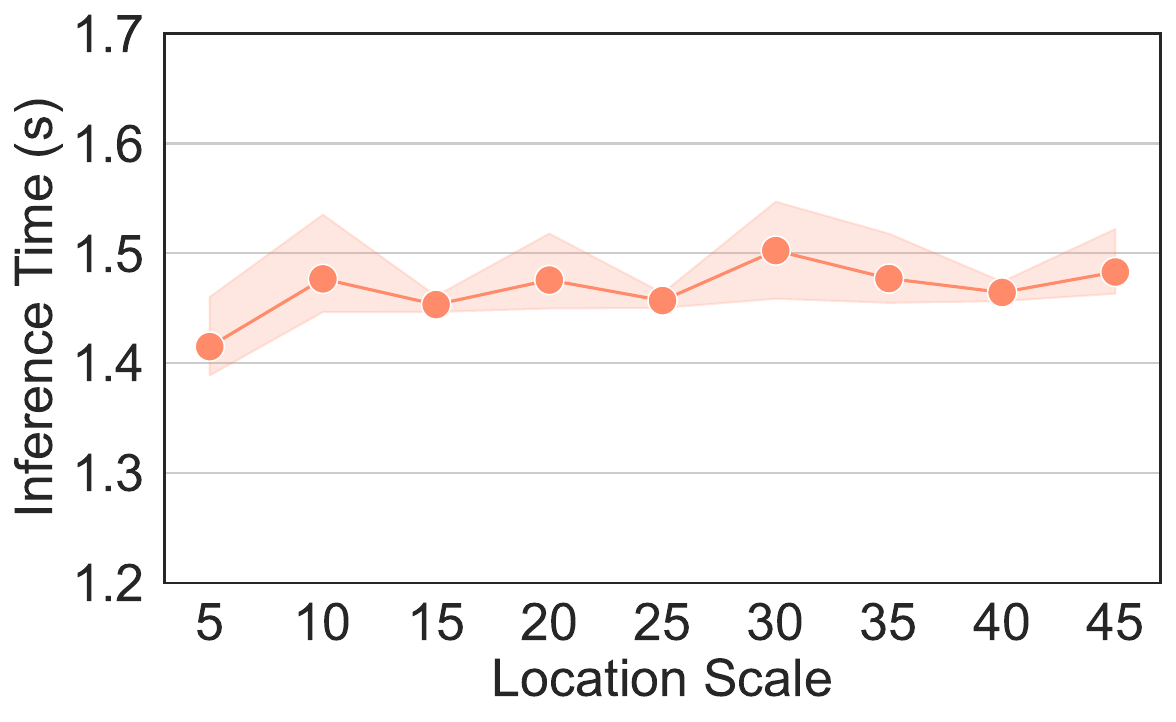}
\vspace{-2mm}
\caption{Inference Analysis with different scale locations under horizon=5 in US-Regions.\label{fig:inference}}
\vspace{-6mm}
\end{figure}

To verify the inference efficiency of HeatGNN, we conduct an inference analysis of different location scales based on the US-States dataset to evaluate the inference time. The results in Figure \ref{fig:inference} indicate that the inference time of the HeatGNN remains stable with only minor fluctuations under different location scales. This consistent performance demonstrates the robustness and reliability of HeatGNN, further highlighting its scalability and suitability for applications involving dynamic or large-scale epidemic transmission graph datasets.

\begin{table}[!b]
\vspace{-4mm}
\centering
\caption{Comparison of inference time (in second) of different models on four datasets with horizon=5.\label{tab:inference}}
\scalebox{0.85}{
\begin{tabular}{c|cccc}
\hline
             & Japan-Prefectures & US-Region & US-State & Australia-COVID \\
             \hline
Cola-GNN     & 0.650 & 0.360  & 0.720 & 0.735 \\
Epi-Cola-GNN & 1.029 & 1.108  & 1.060 & 0.236 \\
HeatGNN      & 0.370 & 0.231  & 0.241 & 0.265 \\
\hline
\end{tabular}
}
\end{table}

\subsection{Robustness Analysis (RQ6)}
Unlike purely data-driven methods, HeatGNN integrates an epidemic model and uses PINN to embed epidemic dynamics into learning. Ensuring consistency with known transmission mechanisms reduces reliance on noisy or incomplete data, improving robustness. We further test with different noisy node proportions by adding Gaussian noise and different missing node proportions by removing nodes, where the results are shown in Table \ref{tab:combined_robustness}. The results indicate that HeatGNN outperforms Cola-GNN and EpiGNN in most cases under different noise and missing node ratios. The results indicate HeatGNN's robustness and prediction stability under incomplete graph structures, and it can more effectively support epidemic transmission prediction tasks in complex public health scenarios.

\begin{table}[!t]
\centering
\caption{Performance under different noisy and missing node proportions in the US-Region with horizon=2.\label{tab:combined_robustness}}
\scalebox{0.75}{
\begin{tabular}{c|c|cccc|cccc}
\hline
 & & \multicolumn{4}{c|}{RMSE $\times 10^3$ ($\downarrow$)} & \multicolumn{4}{c}{PCC ($\uparrow$)} \\
\hline
Perturbation & Proportion & 0\% & 10\% & 20\% & 30\% & 0\% & 10\% & 20\% & 30\% \\
\hline
\multirow{3}{*}{Noisy} 
& Cola-GNN & 0.552 & 0.572 & 0.601 & 0.615 & 0.939 & 0.933 & 0.925 & 0.920 \\
& EpiGNN   & 0.563 & 0.584 & 0.619 & 0.634 & 0.937 & 0.930 & 0.920 & 0.915 \\
& HeatGNN  & 0.541 & 0.561 & 0.597 & 0.615 & 0.941 & 0.935 & 0.926 & 0.920 \\
\hline
\multirow{3}{*}{Missing} 
& Cola-GNN & 0.552 & 0.906 & 1.155 & 1.326 & 0.939 & 0.827 & 0.726 & 0.648 \\
& EpiGNN   & 0.563 & 0.947 & 1.167 & 1.327 & 0.937 & 0.808 & 0.714 & 0.636 \\
& HeatGNN  & 0.541 & 0.906 & 1.094 & 1.247 & 0.941 & 0.822 & 0.742 & 0.669 \\
\hline
\end{tabular}
}
\vspace{-6mm}
\end{table}

\section{Conclusion}
In this study, we propose a novel epidemic forecasting framework, HeatGNN, to improve the forecasting performance. We incorporate the epidemic mechanism model into the epidemiology-informed embedding learning to learn intrinsic epidemic transmission patterns. We use the physics loss to constrain the epidemiology-informed location embeddings to align with real-world epidemics. We propose a mechanistic affinity graph to represent the mechanistic dependence of intrinsic epidemic transmission between locations. We design a heterogeneous transmission graph network to account for the mechanistic heterogeneity. Finally, the forecasting performance is further improved by combining the spatio-temporal graph learning and transmission graph modules. The effectiveness of HeatGNN is validated using real-world datasets for epidemic forecasting. Experimental results demonstrate that HeatGNN is capable of outperforming existing state-of-the-art models in epidemic forecasting. In the future, we plan to extend HeatGNN to different epidemics and construct a more interpretable mechanistic affinity graph with mechanisms models to cope further with the spatio-temporal and mechanistic heterogeneity in epidemic transmission.

 
\bibliography{reference}
\bibliographystyle{IEEEtran}

\begin{IEEEbiography}
[{\includegraphics[width=1in,height=1.25in,clip,keepaspectratio]{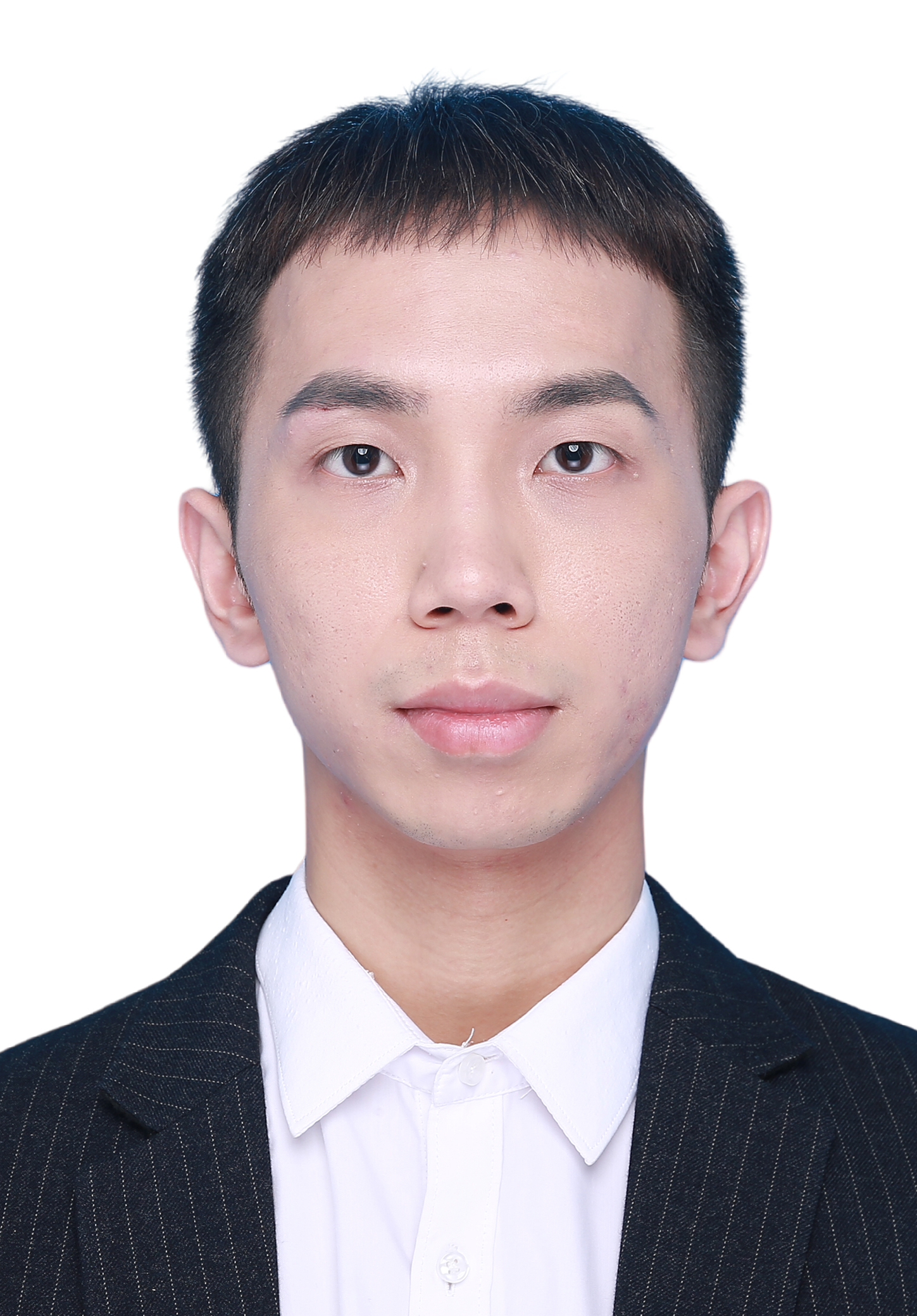}}]{Yufan Zheng}
received the B.S. degree from the Nanfang College of Sun Yat-sen University, Guangdong, China. He was a research assistant at the City University of Hong Kong from December 2022 to July 2024. He is a Ph.D. student at the University of Canberra, Australia, and the ARC Training Centre in Plant Biosecurity. His research interests include data-driven modeling, data mining, and computational epidemiology.
\vspace{-12mm}
\end{IEEEbiography}

\begin{IEEEbiography}[{\includegraphics[width=1in,height=1.25in,clip,keepaspectratio]{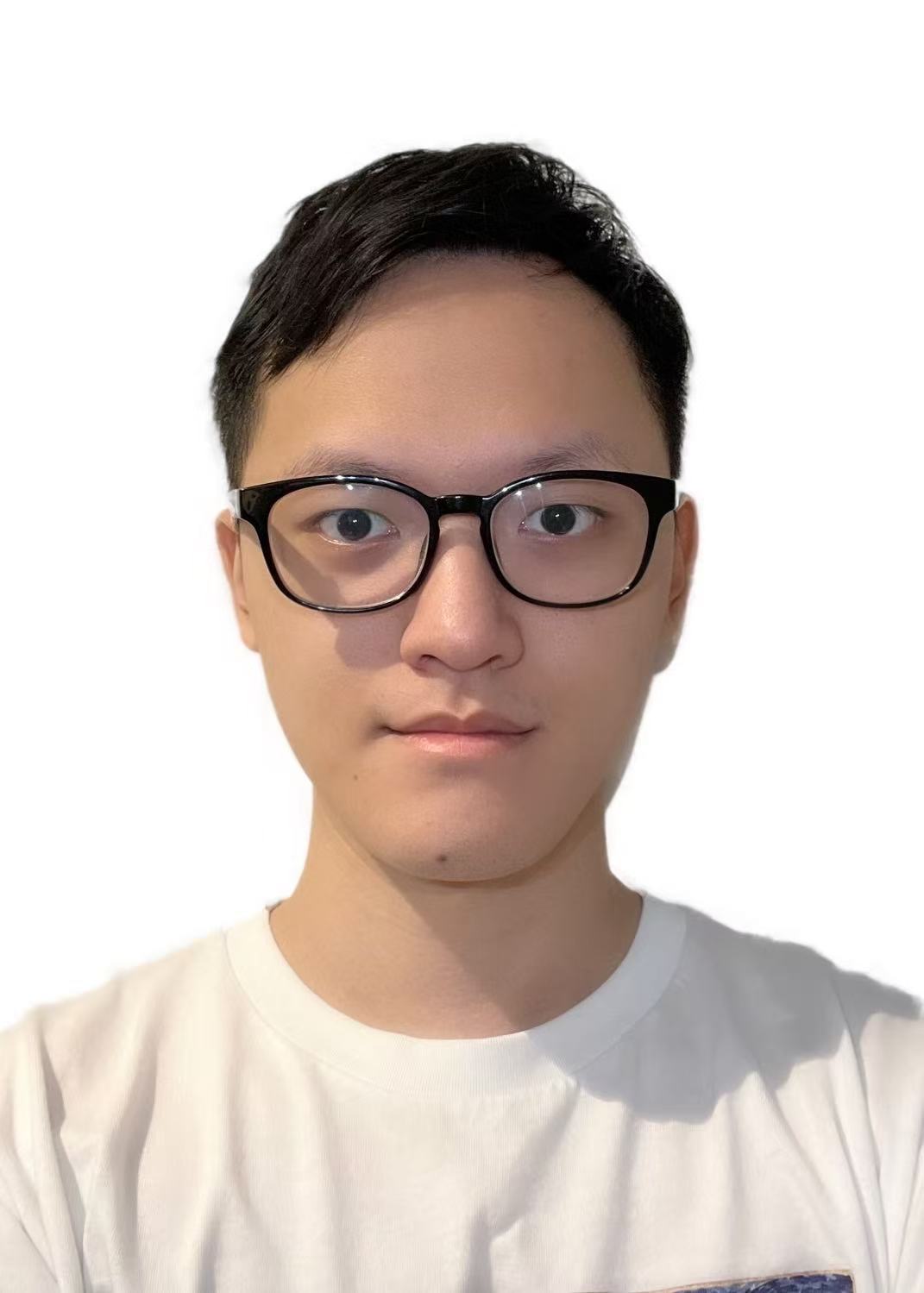}}]{Wei Jiang}
received the Master of Data Science degree from the University of Queensland. He is a Ph.D. student in the School of Electrical Engineering and Computer Science at the University of Queensland. His research interests include social media data mining, graph representation learning, and the applications of large language models.
\vspace{-12mm}
\end{IEEEbiography}

\begin{IEEEbiography}[{\includegraphics[width=1in,height=1.25in,clip,keepaspectratio]{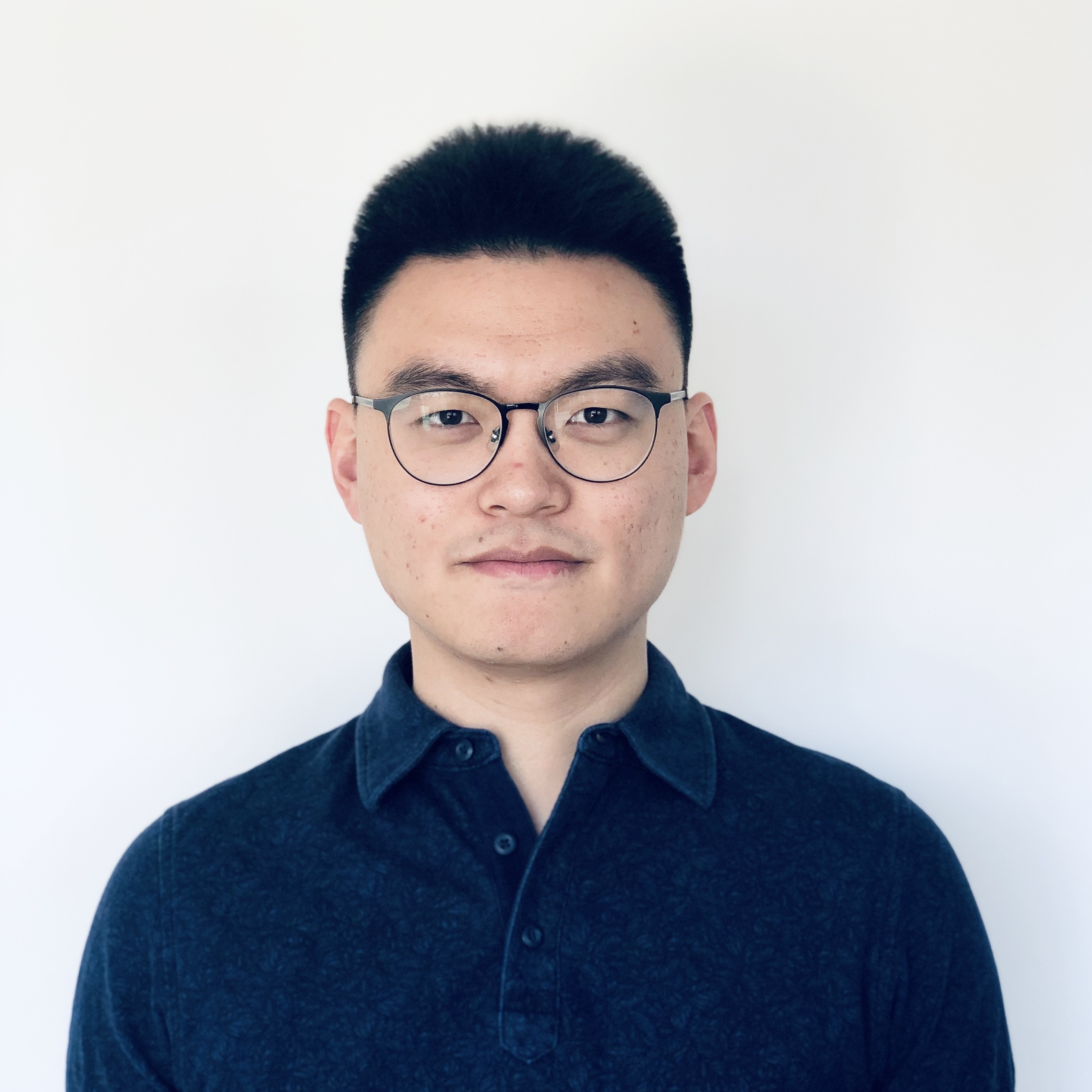}}]{Tong Chen}
received the Ph.D. degree in computer
science from The University of Queensland, Brisbane,
QLD, Australia, in 2020. He is currently a Senior
Lecturer with the Data Science Research Group,
School of Electrical Engineering and Computer Science,
The University of Queensland. His research
interests include data mining, recommender systems,
user behavior modeling, and predictive analytics.
\vspace{-12mm}
\end{IEEEbiography}

\begin{IEEEbiography}[{\includegraphics[width=1in,height=1.25in,clip,keepaspectratio]{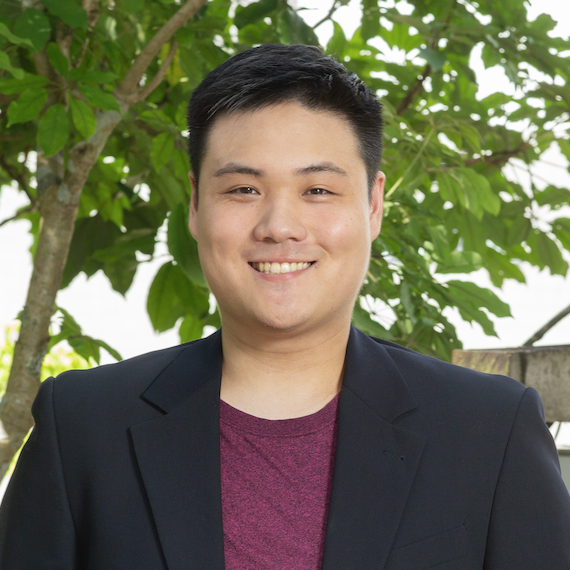}}]{Alexander Zhou}
is currently a Research Assistant Professor in the Department of Computing at The Hong Kong Polytechnic University. He received his PhD from The Hong Kong University of Science and Technology in 2024. His research interests include graph databases and algorithms as well as social science-driven research on networks.
\vspace{-12mm}
\end{IEEEbiography}

\begin{IEEEbiography}[{\includegraphics[width=1in,height=1.25in,clip,keepaspectratio]{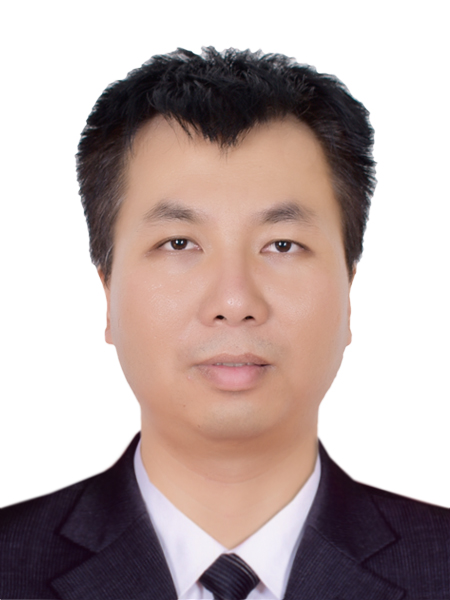}}]{Choujun Zhan}
received the B.S. degree in automatic control engineering from Sun Yat-sen University, Guangzhou, China, in 2007 and the Ph.D. degree in electronic engineering from the City University of Hong Kong in 2012. He worked as a Postdoctoral Fellow at the Hong Kong Polytechnic University after graduation. Since 2016, he has been an Associate Professor at South China Normal University. His research interests include complex networks, time series modeling, epidemic spreading, information diffusion, and machine learning.
\vspace{-10mm}
\end{IEEEbiography}

\begin{IEEEbiography}[{\includegraphics[width=1in,height=1.25in,clip,keepaspectratio]{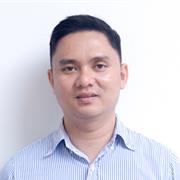}}]{Quoc Viet Hung Nguyen} received the Ph.D. degree from EPFL, Switzerland. He is an associate professor at Griffith University, Australia. His research focuses on data integration, data quality, information retrieval, trust management, recommender systems, machine learning, and big data visualisation. He published several papers in top-tier venues such as SIGMOD, VLDB, WSDM, WWW, SIGIR, KDD, AAAI, ICDE, IJCAI, VLDBJ, IEEE TKDE, and ACM TOIS.
\vspace{-12mm}
\end{IEEEbiography}

\begin{IEEEbiography}[{\includegraphics[width=1in,height=1.25in,clip,keepaspectratio]{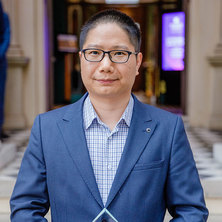}}]{Hongzhi Yin} received the Ph.D. degree in computer science from Peking University, in 2014. He works as an ARC Future Fellow and Full Professor at The University of Queensland, Australia. He has made notable contributions to
recommendation systems, graph learning, and decentralized and edge intelligence. He has published 300+ papers with an H-index of 84 and received the ARC Future Fellowship 2021 and DECRA Award 2016.
\vspace{-12mm}
\end{IEEEbiography}



\vfill

\end{document}